\documentclass[11pt]{article}

\usepackage[final]{acl}

\usepackage{times}
\usepackage{latexsym}

\usepackage[T1]{fontenc}
\usepackage[utf8]{inputenc}

\usepackage{microtype}

\usepackage{inconsolata}

\usepackage{graphicx}
\usepackage{subcaption}
\usepackage{booktabs}
\usepackage{makecell}
\usepackage{multirow}
\usepackage{adjustbox}
\usepackage{amsmath}
\usepackage[table]{xcolor}
\usepackage{tabularx}
\newcommand{\best}[1]{\cellcolor{cyan!20}{#1}}
\newcommand{\worst}[1]{\cellcolor{red!20}{#1}}
\title{Language of Thought Shapes Output Diversity in Large Language Models}

\author{
Shaoyang Xu, \;
Wenxuan Zhang\thanks{Corresponding author} \\ iNLP Lab, Singapore University of Technology and Design \\
\texttt{shaoyang\_xu@mymail.sutd.edu.sg}, \;
\texttt{wxzhang@sutd.edu.sg}
}

\begin{document}
\maketitle
\begin{abstract}

Output diversity is crucial for Large Language Models as it underpins pluralism and creativity.
In this work, we reveal that controlling the language used during model thinking—the \textit{language of thought}—provides a novel and structural source of output diversity.
Our preliminary study shows that different thinking languages occupy distinct regions in a model's thinking space.
Based on this observation, we study two repeated sampling strategies under multilingual thinking—\textit{Single-Language Sampling} and \textit{Mixed-Language Sampling}—and conduct diversity evaluation on outputs that are controlled to be in English, regardless of the thinking language used.
Across extensive experiments, we demonstrate that switching the thinking language from English to non-English languages consistently increases output diversity, with a clear and consistent positive correlation such that languages farther from English in the thinking space yield larger gains.
We further show that aggregating samples across multiple thinking languages yields additional improvements through compositional effects, and that scaling sampling with linguistic heterogeneity expands the model's diversity ceiling.
Finally, we show that these findings translate into practical benefits in pluralistic alignment scenarios, leading to broader coverage of cultural knowledge and value orientations in LLM outputs. Our code is publicly available at \url{https://github.com/iNLP-Lab/Multilingual-LoT-Diversity}.

\end{abstract}

\section{Introduction}

Large Language Models (LLMs) have been globally adopted due to their extensive knowledge and strong reasoning capabilities.
Beyond the correctness of individual responses, this widespread use has drawn increasing attention to the \textit{diversity} of LLM-generated outputs.
Formally, output diversity quantifies a model’s ability to generate multiple distinct responses to open-ended questions without ground-truth answers~\citep{NIPS, chat1}.
It is recognized as a fundamental objective in pluralistic alignment research~\citep{plural,plural2}, where low diversity can lead to homogenization—often referred to as mode collapse~\citep{NIPS, chat1,chat2}—and the over-representation of dominant cultural values~\citep{culture1,culture2}. 
Moreover, diversity is a key indicator of whether AI systems exhibit human-like creativity~\citep{traditional2}, laying the foundation for innovative problem-solving~\citep{math1,math2,math3,math4}, open-ended exploration, and the generation of novel ideas~\citep{idea1,idea2}.

To improve output diversity, temperature scaling is commonly utilized by increasing sampling randomness~\citep{traditional2,temp1,influence1}. 
Other work explored advanced decoding methods~\citep{decoding}, aggregating outputs from multiple LLMs~\citep{multi1,multi2,decoding2}, or increasing prompt variation~\citep{multi2,chat2,multi3}. 
At training time, several studies proposed diversity-driven RLHF and SFT objectives to encourage more varied generations~\citep{influence3,influence4}. 

Despite their promise, most existing work focuses on English-only or multilingual input settings~\citep{multi3}.
In contrast, {leveraging the inherently multilingual nature of modern LLMs~\citep{babel, seallm, qwen3, inclusive}}, we investigate whether the language used during intermediate thinking—referred to as the \textit{language of thought}—can serve as a controllable and structural source of output diversity. Our investigation is motivated by two observations. 
First, insights from cognitive science suggest that multilingualism promotes divergent thinking and creativity, as different languages encode distinct conceptual and structural biases~\citep{theory1,theory2}. 
According to the Sapir--Whorf hypothesis~\citep{theory3}, language can shape how concepts are organized and related during thinking. 
Second, recent studies have demonstrated that modern LLMs are capable of explicit reasoning in multiple languages, with performance differences across languages~\citep{mtts2,mtts5}. 
Together, these insights motivate us to study \textit{language of thought} as a structural property of the model’s thinking process, and to examine how varying this property influences output diversity.

To this end, we begin with a preliminary study that explores \emph{whether different thinking languages induce structural differences in the model’s thinking space} (§\ref{sec:pre}).
Specifically, given the same English input, we control the thinking process to be conducted in different languages and collect the resulting hidden representations.
By visualizing these multilingual thinking representations, we observe that different languages correspond to distinct regions in the model’s thinking space.
Moreover, non-English languages exhibit substantial variation in their distances to English thinking.
These observations reveal geometric differences induced by different languages of thought.

Building on these observations, we next examine \emph{whether the thinking-space shifts induced by different languages of thought help output diversity} (§\ref{sec:sample}\&\ref{sec:exp}). Although the thinking process is controlled to be conducted in different languages, we further control the model’s final outputs to English for fair output diversity evaluation (§\ref{sec:o_c}).
Based on this setup, we perform \textit{repeated sampling} and aggregate the resulting English outputs for diversity evaluation.
Specifically, we explore two sampling strategies.
The first, \textit{Single-Language Sampling}, performs repeated sampling within a single thinking language (§\ref{sec:s_s}).
The second, \textit{Mixed-Language Sampling}, aggregates English outputs generated through thinking in different languages (§\ref{sec:m_s}).

We conduct experiments on two benchmarks using two different diversity metrics.
Multiple LLMs and 15 thinking languages are evaluated (§\ref{sec:setting}).
Our main findings are as follows.

\textbf{First}, under \textit{Single-Language Sampling}, we observe that simply switching the language of thought from English to non-English languages consistently leads to higher output diversity.
By further computing the correlation between output diversity and the thinking-space distance to English across non-English languages, we identify a clear positive relationship:
thinking languages that are geometrically farther from English consistently achieve higher output diversity.
These results demonstrate that sampling within thinking regions outside the English-dominant space can systematically mitigate output homogenization.
We also evaluate output quality and find that thinking in non-English languages incurs only negligible degradation (§\ref{sec:r_s_s}).

\textbf{Second}, we further find that \textit{Mixed-Language Sampling} yields additional gains in output diversity.
This result indicates that sampling from distinct thinking regions induced by linguistic heterogeneity
can further enhance output diversity beyond a single region.
Further analysis reveals clear compositional effects among languages:
while removing any single language has a relatively small impact,
removing multiple languages leads to substantially greater degradation in diversity (§\ref{sec:r_m_s}).

\textbf{Third}, we analyze the effects of the sampling number and temperature, and find that \textit{Mixed-Language Sampling} exhibits a pronounced advantage over \textit{Single-Language Sampling} when further scaling the sampling number,
highlighting the role of linguistic heterogeneity in expanding the model’s diversity ceiling (§\ref{sec:analysis}).

\textbf{Finally}, we extend our analysis to pluralistic alignment scenarios (§\ref{sec:application}).
Our results show that \textit{Mixed-Language Sampling} leads to broader coverage of cultural knowledge and values in LLMs,
outperforming other sampling strategies, including English sampling, high-temperature decoding, explicit diversity requests, and multilingual prompting.
These results highlight the practical utility of our findings in real-world applications.

Overall, our findings establish the \textit{language of thought} as a novel and effective control axis for enhancing output diversity.

\section{Related Work}

\paragraph{Output Diversity of LLMs}
Many studies have shown that LLMs often exhibit limited output diversity~\citep{eval1,eval7,eval8,eval9}.
Output diversity evaluation typically considers lexical, syntactic, and semantic dimensions~\citep{eval2,traditional1,chat2}, and employs tools such as Self-BLEU~\citep{eval6} and Sentence-BERT~\citep{eval5} to compute diversity metrics in NLG tasks~\citep{eval2}.
Moreover, diversity is often evaluated alongside novelty and creativity in more complex generation settings~\citep{chat1,chat2,traditional2,math1,math2}.
Recently, NOVELTYBENCH~\citep{chat1} and INFINITY-CHAT~\citep{NIPS} were introduced to assess the ability of LLMs to produce distinct outputs in open-domain dialogue.

Existing approaches to improve output diversity include aggregating outputs from multiple LLMs~\citep{multi1,multi2}, increasing prompt variation~\citep{multi1,chat2,multi3}, and developing diversity-driven RLHF and SFT objectives~\citep{influence3,influence4}.
Unlike these approaches, our work explores the inherent multilingual properties of LLMs as a structural source of output diversity.

\paragraph{Multilingual Reasoning}
Modern LLMs are trained to perform explicit intermediate reasoning before producing final answers~\citep{tts1,tts2,tts3}. {As LLMs increasingly exhibit a shared—yet still imbalanced—thought space across different languages~\citep{m-understand1, m-understand2, m-understand3}, many studies have explored the multilingual generalization of LLM reasoning~\citep{mtts1,mtts2,mtts3,mtts4,mtts5,mtts6,mtts7}.
Other work has investigated whether multilingualism can improve the performance~\citep{mtts9,mtts10,m-help} and efficiency~\citep{mtts11,mtts12} of reasoning.
However, none of these studies have examined whether multilingual thinking can enhance the output diversity of LLMs.

\section{Language Geometry of Thinking Space}
\label{sec:pre}

We first conduct a preliminary study to examine \emph{whether different thinking languages induce structural differences in the model’s thinking space.}

\subsection{Thinking Language Control}
\label{sec:lang_control}

All of our investigations focus on reasoning-capable LLMs.
Given an English input prompt, the model first performs intermediate thinking $T$,
enclosed within \texttt{<think>...</think>},
and then generates the final output $o$, both in English by default.

To control the LLM to perform its intermediate thinking in a target language $l$,
we follow existing multilingual reasoning techniques~\citep{mtts2,mtts5}.
Specifically, we insert a short prefix,
\texttt{``Okay, the user is asking''}—translated into $l$—
immediately after the \texttt{<think>} token,
guiding the subsequent thinking process to be conducted in the target language.
The translated prefixes, together with a sanity check of the language control,
are provided in Appendix~\ref{sec:lang_control_details}.

\subsection{Visualizing Multilingual Thinking Space}
\label{sec:dis}

\paragraph{Collecting Hidden States}
Given a set of English input questions, we apply thinking language control to encourage the model to perform thinking in language $l$ for each sample.
For a single sample, let the thinking process consist of $N$ tokens $\{t^{(l)}_i\}_{i=1}^N$, and let $h^{(l)}_{i,j}$ denote the hidden state of token $t^{(l)}_i$ at layer $j$.
To obtain a compact representation of the model’s thinking behavior, we first average hidden states across all thinking tokens within a sample, and then further average across all samples.
This yields a single vector representation $h^{(l)}_j$ that summarizes the model’s thinking behavior in language $l$ at layer $j$.
Repeating this process for all thinking languages produces a set of language-specific thinking representations at each layer.

\paragraph{PCA Visualization}
To visualize the geometry of multilingual thinking space, we first normalize all language representations using $\ell_2$ normalization.
Viewing English as the anchor, we then compute the cosine distance between each non-English language $l$ and English at layer $j$ as
$d_j(l,\text{en}) = 1 - \cos\!\left(h^{(l)}_j,\, h^{(\text{en})}_j\right)$.
Finally, we apply PCA to the centered representations to obtain a two-dimensional layout for visualization.
In the resulting plot, PCA determines only the angular arrangement of languages,
while the radial distance of each point is explicitly fixed to its cosine distance to English, i.e., $d_j(l,\text{en})$.

\begin{figure}[t]
    \centering
    \includegraphics[width=1.0\linewidth]{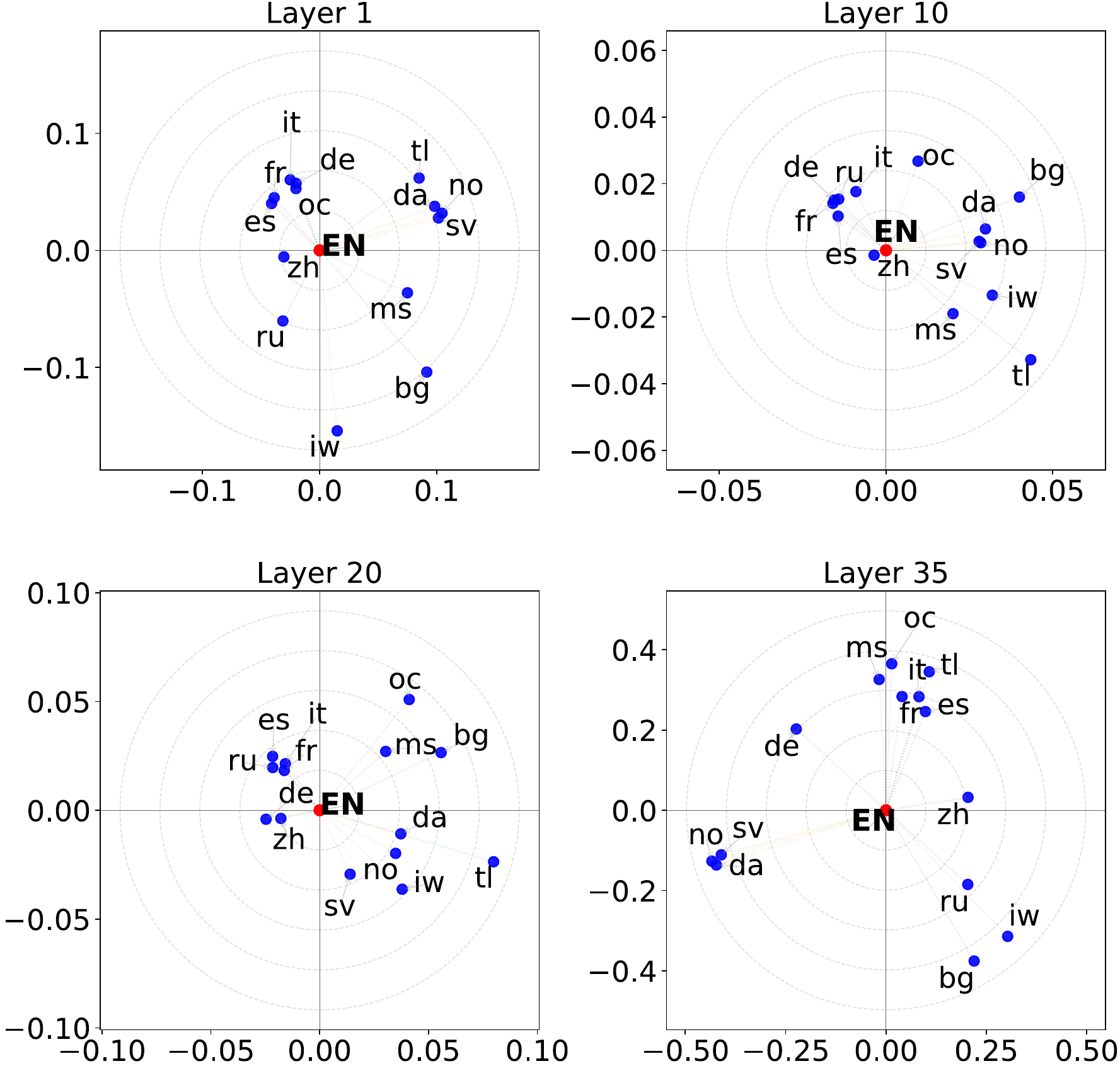}
    \caption{Language geometry of thinking space on Qwen3-8B, with different distance scales across layers for visualization purposes.}
    \label{fig:geometry}
\end{figure}

\subsection{Observations}
\label{sec:obs}

We select 14 non-English languages together with English that are officially supported by Qwen3-8B~\citep{qwen3} to analyze the multilingual thinking space of the model.
Figure~\ref{fig:geometry} shows the resulting geometry at several representative model layers.

\paragraph{Geometric Separation across Thinking Languages}
We first observe clear geometric separation among thinking representations induced by different thinking languages:
representations corresponding to different languages tend to occupy separable regions in the model’s thinking space.
This separation holds consistently across model layers, including intermediate layers that are often assumed to be relatively abstract and less language-specific~\citep{mbert}.
These observations indicate the presence of language-correlated geometric structure in the model’s thinking space.

\paragraph{Varied Distances to English Thinking}
We further observe systematic variation in the geometric distance between non-English languages and English.
Some languages (e.g., \texttt{zh}, \texttt{fr}, \texttt{es}, \texttt{de}) consistently appear closer to English,
whereas others (e.g., \texttt{iw}, \texttt{bg}, \texttt{tl}) are embedded farther away.
Overall, these results indicate that different languages of thought occupy distinct regions of the model’s thinking space,
with varied distances to English.

\section{Repeated Sampling under Multilingual Thinking}
\label{sec:sample}

In this and following sections, we further investigate \emph{whether the thinking-space shifts induced by different languages of thought translate into greater output diversity}.
In this section, we first introduce a controlled output setting and two repeated sampling strategies.
The resulting outputs are used for diversity evaluation in Section~\ref{sec:exp}.

\subsection{Output Language Control}
\label{sec:o_c}

Although the model’s intermediate thinking $T$ is controlled to be conducted in a specific language and enclosed within \texttt{<think>...</think>} (Section~\ref{sec:lang_control}),
we further constrain the final output $o$ to English to enable fair output diversity evaluation.
This is achieved by inserting an additional English prefix immediately after \texttt{</think>}—\texttt{Let me provide my answer in English only:}—
to guide the model to generate the final response in English.
Only the English final outputs are collected for subsequent output diversity evaluation.

Appendix~\ref{sec:lang_control_details} provides a sanity check indicating that both the thinking and output segments largely follow the intended language control.

\subsection{Single-Language Sampling}
\label{sec:s_s}

Section~\ref{sec:obs} shows that different non-English languages occupy distinct thinking regions with varying distances from English.
This motivates us to examine \emph{whether switching to a thinking region away from English and performing repeated sampling within that region leads to increased output diversity.}
To this end, we introduce the first repeated sampling strategy, \textit{Single-Language Sampling}.

Given an English input, the model’s intermediate thinking is constrained to a fixed thinking language $l$, while the final output is generated in English.
We then sample the model $M$ times under this fixed thinking language, and aggregate the resulting English outputs into a set $\mathcal{O}_l$ for diversity evaluation.

\subsection{Mixed-Language Sampling}
\label{sec:m_s}

We further examine \emph{whether sampling from distinct thinking regions induced by different languages can yield additional gains in output diversity}.
This setting allows us to investigate the compositional effects of multiple thinking languages on output diversity.
We thus introduce our second repeated sampling strategy, \textit{Mixed-Language Sampling}.

Specifically, given an English input, we sample the model $M$ times, each time controlling the model to perform intermediate thinking in a different language, while keeping the final output in English.
The resulting outputs are aggregated into a set of outputs $\mathcal{O}_{\text{mixed}}$, on which the same diversity evaluation is conducted.

\begin{table*}[t]
\centering
\small
\begin{adjustbox}{max width=\textwidth}
\begin{tabular}{lc|cccccccccccccc|c}
\toprule
 & en & it & ms & zh & ru & de & iw & bg & da & no & sv & es & tl & oc & fr & avg (non-en) \\
\midrule
\multicolumn{17}{c}{\textit{Distinct Score $\uparrow$}} \\
\midrule
Qwen3-8B &
\worst{28.55} & 34.60 & 33.47 & {29.00} & 34.14 & 35.67 &
\best{41.33} & {39.80} & 36.03 & {39.69} & 36.73 & {32.33} &
38.35 & 38.87 & 33.93 & 36.00 \\

Qwen3-14B &
\worst{26.20} & 30.67 & 29.23 & {28.80} & 31.40 & {28.93} &
\best{36.87} & 32.13 & 30.13 & {34.55} & 32.33 & 29.73 &
32.68 & {33.26} & 29.53 & 31.45 \\

Qwen3-32B &
\worst{35.00} & 39.33 & {37.78} & {37.80} & 38.67 & 39.73 &
\best{43.38} & 39.93 & 40.67 & 40.22 & {41.80} & 39.73 &
41.41 & {42.96} & 40.80 & 40.30 \\

DeepSeek-14B &
{38.33} & 43.47 & \worst{38.07} & 41.33 & 44.60 & 41.14 &
49.63 & 47.13 & {51.85} & {52.40} & 50.60 & 43.60 &
\best{52.42} & 45.93 & {42.27} & 46.03 \\

\midrule
\multicolumn{17}{c}{\textit{Similarity Score $\downarrow$}} \\
\midrule
Qwen3-8B &
\worst{87.28} & 85.43 & {86.53} & {86.73} & 85.57 & 85.14 &
{83.66} & 84.89 & 84.79 & 83.93 & 85.14 & 85.76 &
{83.20} & \best{80.79} & 84.57 & 84.72 \\

Qwen3-14B &
\worst{87.82} & 86.68 & 87.30 & 86.89 & 87.20 & {87.78} &
\best{85.04} & 86.94 & 86.81 & {86.17} & 86.46 & 87.35 &
{87.36} & {85.72} & 87.19 & 86.78 \\

Qwen3-32B &
\worst{82.10} & 80.59 & {81.76} & {81.61} & 80.67 & {78.00} &
79.64 & 81.45 & 79.78 & 79.54 & {79.06} & 79.84 &
79.71 & \best{77.65} & 80.62 & 79.99 \\

DeepSeek-14B &
81.15 & 79.98 & \worst{83.28} & {82.11} & 80.17 & 81.08 &
\best{76.16} & 81.34 & {77.56} & 77.61 & 79.27 & 81.12 &
{76.70} & 79.81 & {81.88} & 79.86 \\

\midrule
\multicolumn{17}{c}{\textit{Output Quality $\uparrow$}} \\
\midrule
Qwen3-8B &
\best{96.82} & 95.86 & 95.72 & 95.53 & 96.11 & {96.69} &
95.53 & 96.04 & {95.09} & \worst{95.00} & \best{96.82} & 95.72 &
95.70 & 95.59 & {95.40} & 95.80 \\

Qwen3-14B &
\best{96.93} & {94.94} & 95.48 & {95.03} & \worst{94.70} & 96.03 &
96.50 & 96.00 & 96.10 & {96.78} & {96.16} & 95.79 &
95.49 & 95.87 & 95.75 & 95.80 \\

Qwen3-32B &
\best{97.36} & 96.08 & 95.85 & 96.22 & 95.36 & {94.47} &
95.57 & {97.07} & 95.52 & {96.87} & 95.96 & {94.97} &
96.04 & 96.19 & \worst{94.26} & 95.70 \\

DeepSeek-14B &
\best{95.84} & 94.75 & 93.94 & 94.71 & 93.69 & 93.27 &
\worst{89.17} & 94.52 & 92.95 & {92.60} & 93.66 & {94.93} &
{90.73} & {95.45} & {95.80} & 93.60 \\

\bottomrule
\end{tabular}
\end{adjustbox}
\caption{
Distinct Score (\%), Similarity Score (\%), and Output Quality across models and thinking languages under \textit{Single-Language Sampling} on \textsc{NoveltyBench}.
For each row, the best and worst language results are highlighted.
}
\label{tab:diversity_main}
\end{table*}

\section{How Does Language of Thought Shape Output Diversity?}
\label{sec:exp}
\subsection{Experiment Settings}
\label{sec:setting}

\paragraph{Datasets and Evaluation Metrics}
We evaluate output diversity on two benchmarks, \textsc{NoveltyBench}~\citep{chat1} and \textsc{Infinity-Chat}~\citep{NIPS}, each containing 100 open-ended questions without ground-truth answers.
Given an input question, we sample the model $M$ times to obtain a set of outputs $\mathcal{O}$ and evaluate their diversity and quality.
Following the evaluation protocols of the original datasets, we consider two output diversity metrics and one output quality metric, as described below.

\textbf{Metric 1: Distinct Score.}
We compute \textit{Distinct Score} to measure the functional distinctiveness of $\mathcal{O}$ following~\citet{chat1}.
Specifically, the \texttt{deberta-v3-large-generation-similarity}\footnote{https://huggingface.co/yimingzhang/deberta-v3-large-generation-similarity} model is used to sequentially judge whether two outputs are functionally equivalent.
Each output $o_i$ is compared with all previous outputs $\{o_1, \dots, o_{i-1}\}$.
If $o_i$ is judged equivalent to any $o_j$ ($j < i$), it is assigned to the same equivalence class; otherwise, it forms a new class.
The $M$ outputs are thus clustered into $C$ equivalence classes, and the \textit{Distinct Score} is defined as $C/M$.

\textbf{Metric 2: Similarity Score.}
We also compute the \textit{Similarity Score} following~\citet{NIPS}, which captures semantic similarity among outputs in $\mathcal{O}$.
Sentence-level embeddings are first obtained for all generated outputs, and cosine similarity is computed for all output pairs.
The final score is obtained by averaging cosine similarities across all pairs.
We use \texttt{Qwen3-Embedding-8B}\footnote{https://huggingface.co/Qwen/Qwen3-Embedding-8B} for embedding extraction.

\textbf{Metric 3: Output Quality.}
To assess whether improvements in output diversity come at the cost of output quality, we evaluate the quality of responses in $\mathcal{O}$ using \texttt{gpt-4o-mini}, with scores ranging from 0 to 100.
The evaluation considers two dimensions: instruction adherence and overall response quality.
Details of the evaluation prompting are provided in Appendix~\ref{sec:quality_evaluation}.

\paragraph{Languages and LLMs}
\label{sec:languages_llms}

We conduct experiments on the thinking mode of the Qwen3 family~\citep{qwen3} with model sizes 8B, 14B, and 32B, as well as DeepSeek-R1-Distill-Qwen-14B (DeepSeek-14B)~\citep{tts3}.
We select 15 thinking languages for evaluation:
\texttt{en}, \texttt{it}, \texttt{ms}, \texttt{zh}, \texttt{ru}, \texttt{de}, \texttt{iw}, \texttt{bg}, \texttt{da}, \texttt{no}, \texttt{sv}, \texttt{es}, \texttt{tl}, \texttt{oc}, and \texttt{fr}, 
from the supported languages of the tested models.

\paragraph{Sampling Parameters}
\label{sec:sampling_params}

Unless otherwise specified, the decoding temperature is set to $0.6$.
For fair comparison across sampling strategies, the number of samples $M$ is set equal to the number of thinking languages, i.e., $M = 15$.

\subsection{Results on Single-Language Sampling}
\label{sec:r_s_s}

\paragraph{Main Diversity Results}
Table~\ref{tab:diversity_main} summarizes the output diversity results on \textsc{NoveltyBench}.
On average, switching the thinking language from English to non-English languages
yields an improvement of 5.3 to 7.7 points in \textit{Distinct Score}
and a reduction of 1.04 to 2.56 points in \textit{Similarity Score}.
These results suggest that sampling from thinking regions outside the English-dominant space
provides a systematic advantage in output diversity.

We also observe substantial variation in output diversity across thinking languages.
Besides \texttt{en}, some languages such as \texttt{ms} and \texttt{zh} consistently exhibit lower diversity,
whereas others, including \texttt{iw}, \texttt{no}, and \texttt{oc}, achieve substantially higher diversity across models and metrics.
In some cases, individual languages lead to particularly large gains.
For example, thinking in \texttt{iw} on Qwen3-8B improves the \textit{Distinct Score} by 12.78 points compared to \texttt{en}.
Taken together with the geometric findings from Section~\ref{sec:obs},
these results highlight the strong potential of specific thinking languages—
especially those farther from English in the thinking space—
for enhancing output diversity.

\begin{figure}[t]
    \centering
    \includegraphics[width=0.893\linewidth]{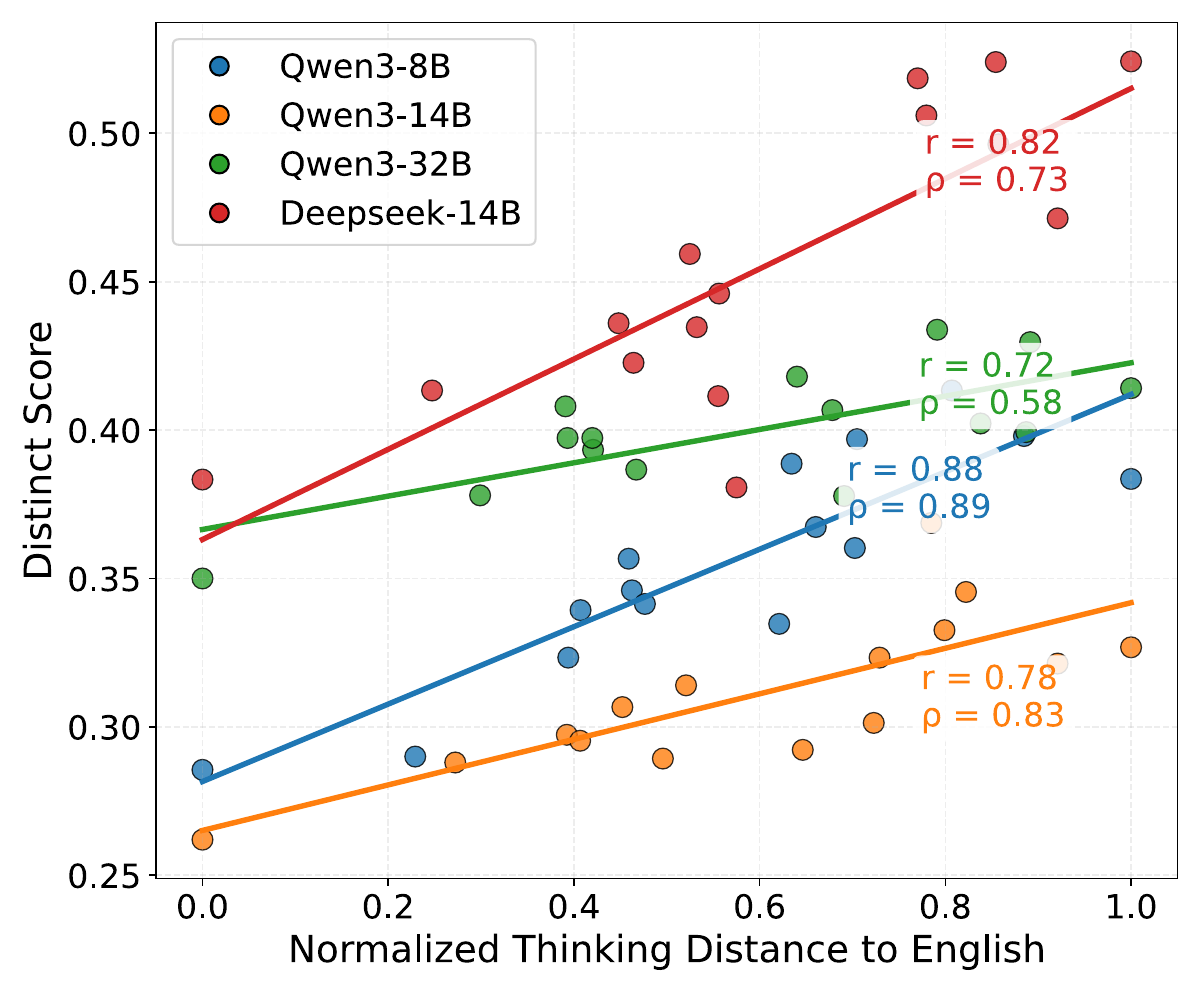}
    \caption{
    Correlation between the Distinct Score and the thinking distance to English across languages.
    Pearson’s $r$ and Spearman’s $\rho$ are reported for each model.
    Distinct Scores are obtained under \textit{Single-Language Sampling} on \textsc{NoveltyBench}.
    Thinking distances are normalized to the range $[0,1]$.
    }
    \label{fig:col}
\end{figure}

\paragraph{Correlation with Thinking Distance to English}

We further examine the relationship between the geometric properties of the thinking space and output diversity.
For each language $l$, we compute its thinking distance to English, $d(l,\text{en})$, by averaging the layer-wise distances $d_j(l,\text{en})$ across all model layers (Section~\ref{sec:dis}), where English has distance zero. To ensure comparability between languages, we normalize the thinking distances to the range of [0, 1]. We then analyze the correlation between these thinking distances and the output diversity under \textit{Single-Language Sampling}.
Figure~\ref{fig:col} reports the Pearson and Spearman correlations on \textsc{NoveltyBench}, with output diversity measured by the \textit{Distinct Score}.

We observe a strong positive correlation across different models, with Pearson’s $r$ ranging from 0.72 to 0.88 and Spearman’s $\rho$ ranging from 0.58 to 0.89.
These results corroborate our earlier observations, indicating that the distance to English in the thinking space is informative of the output diversity achievable under \textit{Single-Language Sampling}.
More specifically, languages that are geometrically farther from English tend to correspond to more distinct thinking regions, and repeated sampling within such regions is associated with higher output diversity.

\paragraph{Output Diversity vs. Quality}
Table~\ref{tab:diversity_main} also reports the output quality results.
We observe a mild trade-off between output diversity and quality.
While English generally achieves higher output quality,
there is no clear pattern in which languages with the highest output diversity consistently suffer the lowest output quality.
In some cases, specific languages such as \texttt{sv} and \texttt{oc} achieve strong performance on both dimensions.
Overall, thinking in non-English languages results in only a modest decrease of 1.02 to 2.24 points in output quality.

Appendix~\ref{sec:res_infi} provides results on \textsc{Infinity-Chat}, which also exhibits similar patterns.

\begin{table}[t]
\centering
\footnotesize
\setlength{\tabcolsep}{5pt}
\renewcommand{\arraystretch}{1.05}
\resizebox{\columnwidth}{!}{%
\begin{tabular}{lcccc}
\toprule
Model & S-en & S-non-en avg & S-best & Mixed \\
\midrule
\multicolumn{5}{c}{\textsc{NoveltyBench}} \\
\midrule
Qwen3-8B   & 28.55 & 36.00 & 41.33 & \textbf{43.73} \\
Qwen3-14B  & 26.20 & 31.45 & 36.87 & \textbf{38.00} \\
Qwen3-32B  & 35.00 & 40.30 & 43.38 & \textbf{46.53} \\
DeepSeek-14B
           & 38.33 & 46.03 & \textbf{52.42} & 52.07 \\
\midrule
\multicolumn{5}{c}{\textsc{Infinity-Eval}} \\
\midrule
Qwen3-8B   & 20.67 & 22.54 & 24.51 & \textbf{28.13} \\
Qwen3-14B  & 20.40 & 22.60 & \textbf{27.07} & 26.73 \\
Qwen3-32B  & 27.00 & 27.52 & 28.66 & \textbf{31.47} \\
DeepSeek-14B
           & 25.27 & 31.84 & \textbf{39.61} & 35.33 \\
\bottomrule
\end{tabular}%
}
\caption{
Distinct score (\%) comparison of \textit{Mixed-Language Sampling} and \textit{Single-Language Sampling} on \textsc{NoveltyBench} and \textsc{Infinity-Chat}. \textbf{Bold} indicates the best-performing sampling setting for each model and benchmark.
}
\label{tab:mixed_vs_single_distinct}
\end{table}

\subsection{Results on Mixed-Language Sampling}
\label{sec:r_m_s}

\paragraph{Comparison with Single-Language Sampling}

Table~\ref{tab:mixed_vs_single_distinct} compares \textit{Mixed-Language Sampling} with three \textit{Single-Language Sampling} settings:
English sampling (S-en), the average performance over non-English sampling (S-non-en avg),
and the best-performing single-language sampling (S-best).
Across both benchmarks, \textit{Mixed-Language Sampling} consistently improves output diversity over S-en and S-non-en avg. 

Moreover, \textit{Mixed-Language Sampling} often matches or even exceeds the performance of the S-best setting.
These results indicate that \textit{Mixed-Language Sampling} provides a robust strategy for improving output diversity
without requiring prior knowledge of which single language performs best.
This advantage arises from the structural differences among languages in the thinking space (Section~\ref{sec:obs}):
sampling from multiple distinct thinking regions and aggregating the resulting outputs exploits the compositional effects of different languages.

Results based on the \textit{Similarity Score} are reported in Appendix~\ref{sec:mix_result} and show the same trend.

\begin{figure}[t]
    \centering
    \includegraphics[width=0.7\linewidth]{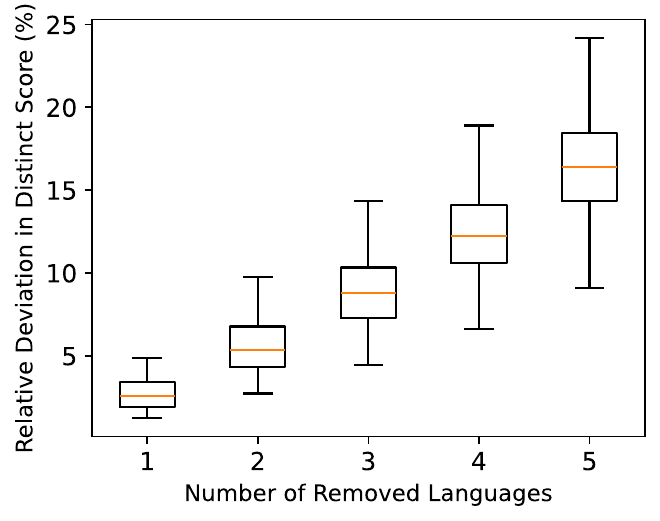}
    \caption{
        Relative deviation in Distinct Score under the removal of $k$ languages
        in \textit{Mixed-Language Sampling}.
    }
    \label{fig:remove_k_boxplot}
\end{figure}

\paragraph{Compositional Effects of Different Languages}
To further explore the compositional effects of different languages in \textit{Mixed-Language Sampling},
we conduct an ablation study on Qwen3-8B by progressively removing $k$ languages
from \textit{Mixed-Language Sampling} ($k = 1,\dots,5$).
For each value of $k$, we enumerate all possible combinations of language removal and measure the relative deviation of the \textit{Distinct Score} from the original result, to quantify the effect of language removal.

Figure~\ref{fig:remove_k_boxplot} shows the relative deviation in \textit{Distinct Score}.
We first observe that removing a single language leads to only a small change
($2.7\%$ on average), indicating that \textit{Mixed-Language Sampling} does not rely on any individual
language to achieve its diversity gains.
However, as $k$ increases, the diversity degradation grows rapidly and in a superlinear manner.
This suggests that the contributions of different languages are not merely additive;
instead, languages provide complementary diversity benefits through their joint participation. Together, these results demonstrate that output diversity under \textit{Mixed-Language Sampling}
emerges from the compositional interaction of multiple languages,
rather than from any single dominant language.

\subsection{Other Analysis}
\label{sec:analysis}

Two parameters are important in repeated sampling:
the sampling number $M$ and the temperature.
By default, we set $M = 15$ and the temperature to $0.6$.
In this section, we vary these parameters using Qwen3-8B to examine their effects on two sampling strategies.
For \textit{Single-Language Sampling}, we select four representative languages for analysis:
\texttt{en} and \texttt{zh} (lower-performing), and \texttt{bg} and \texttt{iw} (higher-performing).

\begin{figure}[t]
    \centering
    \includegraphics[width=1.0\linewidth]{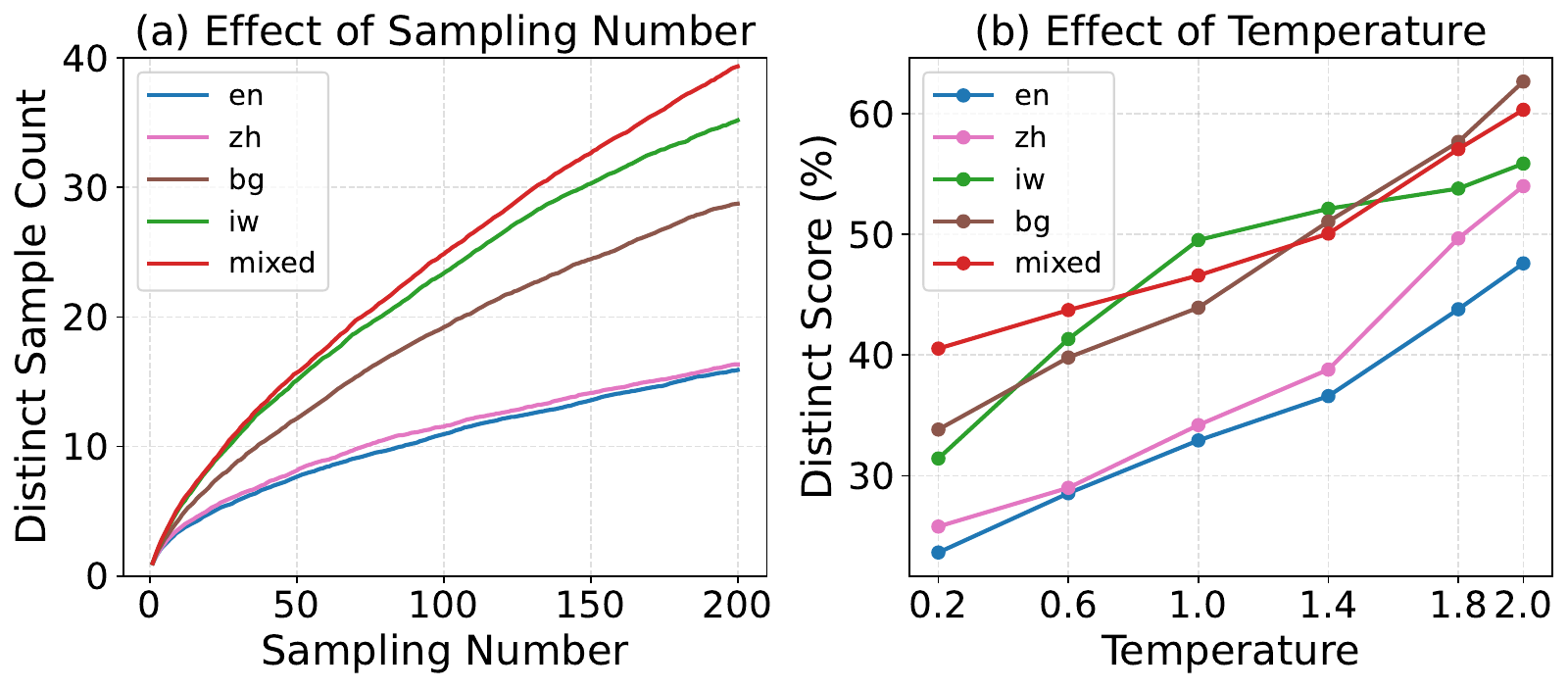}
    \caption{
    Effects of sampling parameters on output diversity.
    (a) Distinct sample count as a function of the sampling number $M$ at a fixed temperature ($0.6$).
    (b) Distinct Score (\%) under different temperatures with a fixed sampling number ($M = 15$).
    }
    \label{fig:sampling_hyperparameters}
\end{figure}

\subsubsection{Scaling Sampling Number}
We first vary the sampling number $M$ from 1 to 200 while keeping the temperature fixed at $0.6$.
For \textit{Mixed-Language Sampling}, we utilize the full language pool supported by Qwen3
(approximately 100 languages) and randomly select one language as the thinking language for each sampling.
Rather than \textit{Distinct Score} $C/M$, Figure~\ref{fig:sampling_hyperparameters}(a)
directly reports the number of distinct samples $C$.

Across all settings, we observe that the growth of $C$ slows down as $M$ increases,
suggesting the existence of an upper bound on achievable output diversity.
However, \textit{Mixed-Language Sampling} exhibits
a much slower saturation rate compared to \textit{Single-Language Sampling}.
As $M$ increases, its advantage over all \textit{Single-Language Sampling} settings continues to widen.

This behavior indicates that \textit{Mixed-Language Sampling} effectively expands the model’s diversity ceiling.
Such an expansion arises from the increased coverage of distinct thinking regions enabled by linguistic heterogeneity.
Although we explore over 100 languages, further unlocking the benefits of linguistic diversity remains an interesting direction for future work.

\subsubsection{Varying Temperatures}
We next fix the sampling number $M$ at 15 and vary the temperature
over $\{0.2, 0.6, 1.0, 1.4, 1.8, 2.0\}$.
The results are shown in Figure~\ref{fig:sampling_hyperparameters}(b).

We observe a compositional effect between our language of thought and temperature scaling:
while switching the language of thought from English to other languages already improves output diversity,
increasing the temperature further yields additional gains.
Moreover, we observe clear advantages of \textit{Single-Language Sampling} with specific non-English languages and \textit{Mixed-Language Sampling}.
For instance, \textit{Mixed-Language Sampling} at temperature $1.0$ achieves diversity comparable to English sampling at temperature $2.0$.

\begin{table}[t]
\centering
\small
\setlength{\tabcolsep}{5pt}
\renewcommand{\arraystretch}{1.05}
\begin{tabular}{cc|cc}
\toprule
\textbf{Model} & \textbf{Method} & \textbf{Blend} & \textbf{WVS} \\
\midrule

\multirow{5}{*}{Qwen3-8B}
& ES  & 67.9        & 40.0 \\
& HT  & 68.0 (+0.1) & 39.0 (-1.0) \\
& RD  & 73.3 (+5.4) & 52.7 (+12.7) \\
& MP  & 76.1 (+9.2) & 52.0 (+12.0) \\
& MLS & \textbf{76.7 (+8.8)} & \textbf{59.0 (+19.0)} \\
\midrule

\multirow{5}{*}{Qwen3-14B}
& ES  & 66.7        & 31.6 \\
& HT  & 67.1 (+0.4) & 32.7 (+1.1) \\
& RD  & 68.4 (+1.7) & 38.0 (+6.4) \\
& MP  & 72.7 (+6.0) & 45.1 (+13.5) \\
& MLS & \textbf{74.0 (+7.3)} & \textbf{48.4 (+16.8)} \\
\midrule

\multirow{5}{*}{Qwen3-32B}
& ES  & 67.5        & 40.1 \\
& HT  & 69.2 (+1.7) & 43.6 (+3.5) \\
& RD  & 72.8 (+5.3) & \textbf{53.4 (+13.3)} \\
& MP  & 73.4 (+5.9) & 46.1 (+6.0) \\
& MLS & \textbf{74.6 (+7.1)} & 50.4 (+10.3) \\
\midrule

\multirow{5}{*}{DeepSeek-8B}
& ES  & 78.6        & 52.3 \\
& HT  & 80.7 (+2.1) & 60.1 (+7.8) \\
& RD  & 78.6 (+0.0) & 54.7 (+2.4) \\
& MP  & 80.6 (+2.0) & 67.2 (+14.9) \\
& MLS & \textbf{83.0 (+4.4)} & \textbf{73.3 (+21.0)} \\
\bottomrule
\end{tabular}

\caption{
Cultural pluralism performance (entropy normalized to 0--100).
Methods: ES (English Sampling), HT (High Temperature), RD (Request Diversity),
MP (Multilingual Prompting), MLS (Mixed-Language Sampling).
Parentheses show absolute gains/losses relative to ES within each model and benchmark.
\textbf{Bold} indicates the best-performing setting per model and benchmark.
}
\label{tab:culture}
\end{table}

\section{Application: Pluralistic Alignment}
\label{sec:application}
In this section, we explore the practical utility of \textit{Mixed-Language Sampling},
given its advantages.
Specifically, we focus on pluralistic alignment scenarios, {where LLM outputs are expected to reflect cultural pluralism~\citep{culture3, plural, culture4, culture5}}.

\subsection{Settings}

\paragraph{Data}
We consider two types of cultural pluralism: \emph{cultural knowledge} and \emph{cultural values},
evaluated using the \textsc{Blend}~\citep{blend} and \textsc{WVS}~\citep{wvs} datasets, respectively.
Both datasets consist of multiple-choice questions.

\paragraph{Evaluation} 
Following~\citet{multi3}, for each cultural question, we perform repeated sampling to obtain $M$ responses
and measure cultural pluralism based on the resulting output distribution.
For \textsc{Blend}, where each option is associated with one or more countries,
we map the sampled outputs to countries and compute the entropy over the country distribution.
For \textsc{WVS}, we directly compute the entropy over the output distribution,
which characterizes the diversity of value orientations reflected in the model responses.

\paragraph{LLMs} 
Experiments are conducted on {Qwen3-8B}, {Qwen3-14B}, {Qwen3-32B},
and {DeepSeek-R1-Distill-Llama-8B} (DeepSeek-8B), with temperature set to 0.6 by default.

\paragraph{Sampling Strategies}
We compare the following sampling strategies:
(1) \textit{English Sampling}, where the language of thought is English;
(2) \textit{High Temperature}, where the temperature is increased to 1.0 while keeping English as the thinking language;
(3) \textit{Request Diversity}, where the model is explicitly instructed to generate novel responses;
(4) \textit{Multilingual Prompting}~\citep{multi3}, where each cultural question is translated into the same 15 languages used in previous experiments;
and (5) \textit{Mixed-Language Sampling}, where the language of thought varies across the same 15 languages used in previous experiments.

The sampling number $M$ is set to 15 for all strategies.
For \textit{Multilingual Prompting} and \textit{Mixed-Language Sampling}, each language is sampled once.

Additional details on the datasets, evaluation protocols, and baselines are provided in Appendix~\ref{sec:culture_evaluation}.

\subsection{Results}

The results in Table~\ref{tab:culture} clearly demonstrate the practical advantage of \textit{Mixed-Language Sampling} for pluralistic alignment.
Across benchmarks and models, \textit{Mixed-Language Sampling} consistently achieves the highest cultural pluralism performance,
enabling LLMs to reflect more diverse cultural knowledge and value orientations.

In contrast, simply increasing the temperature, explicitly requesting diversity, or using multilingual inputs does not yield improvements comparable to \textit{Mixed-Language Sampling}.
These results highlight the practical value of diversifying the language of thought as a means of more fully exploiting the model’s thinking space for pluralistic alignment.

\section{Conclusion}

In this paper, we establish that controlling the \textit{language of thought} provides a structural source of output diversity in LLMs.
We find that switching the thinking language from English to non-English languages consistently increases output diversity,
with stronger gains observed for languages farther from English in the thinking space.
We further demonstrate that aggregating samples across multiple thinking languages yields additional diversity improvements through their compositional effects,
and that scaling the sampling number with linguistic heterogeneity effectively expands the model’s diversity ceiling.
Finally, we show that these findings translate into broader coverage of cultural knowledge and values of LLMs in pluralistic alignment.

\section{Limitations}

This work has two main limitations.

First, while we observe a positive correlation between the geometric distance of non-English thinking languages from English and the output diversity achieved under repeated sampling, there are still several open questions that are not addressed in this work.
For example, many cross-lingual alignment methods explicitly aim to align non-English representations toward English.
An important question is whether such alignment procedures may inadvertently reduce the output diversity associated with aligned non-English languages, and if so, what mechanisms or strategies could mitigate this effect.
Addressing these questions would require controlled interventions on the model, which we leave for future work.

Second, although we demonstrate the practical utility of our findings in pluralistic alignment settings, our evaluation relies on output entropy as a proxy for cultural pluralism.
This experimental setup remains an abstraction of real-world deployment scenarios.
In practice, pluralistic alignment often requires models to align with multiple specific and context-dependent cultural values under explicit constraints.
The sampling strategies studied in this work would likely need to be further adapted—e.g., by incorporating culturally contextualized language-of-thought routing—to be effective in such settings, which we leave for future investigation.

\bibliography{custom}

@inproceedings{inclusive,
  author       = {Wenxuan Zhang},
  editor       = {Sven Koenig and
                  Chad Jenkins and
                  Matthew E. Taylor},
  title        = {Towards Inclusive {AI:} Advancing Multilingual Large Language Models},
  booktitle    = {Fortieth {AAAI} Conference on Artificial Intelligence, Thirty-Eighth
                  Conference on Innovative Applications of Artificial Intelligence,
                  Sixteenth Symposium on Educational Advances in Artificial Intelligence,
                  {AAAI} 2026, Singapore, January 20-27, 2026},
  pages        = {39848},
  publisher    = {{AAAI} Press},
  year         = {2026},
  url          = {https://doi.org/10.1609/aaai.v40i47.41365},
  doi          = {10.1609/AAAI.V40I47.41365},
  bibsource    = {dblp computer science bibliography, https://dblp.org}
}

@article{culture3,
  author       = {Yong Cao and
                  Li Zhou and
                  Seolhwa Lee and
                  Laura Cabello and
                  Min Chen and
                  Daniel Hershcovich},
  title        = {Assessing Cross-Cultural Alignment between ChatGPT and Human Societies:
                  An Empirical Study},
  journal      = {CoRR},
  volume       = {abs/2303.17466},
  year         = {2023},
  url          = {https://doi.org/10.48550/arXiv.2303.17466},
  doi          = {10.48550/ARXIV.2303.17466},
  eprinttype   = {arXiv},
  eprint       = {2303.17466},
  bibsource    = {dblp computer science bibliography, https://dblp.org}
}

@inproceedings{culture4,
  author       = {Shaoyang Xu and
                  Yongqi Leng and
                  Linhao Yu and
                  Deyi Xiong},
  editor       = {Luis Chiruzzo and
                  Alan Ritter and
                  Lu Wang},
  title        = {Self-Pluralising Culture Alignment for Large Language Models},
  booktitle    = {Proceedings of the 2025 Conference of the Nations of the Americas
                  Chapter of the Association for Computational Linguistics: Human Language
                  Technologies, {NAACL} 2025 - Volume 1: Long Papers, Albuquerque, New
                  Mexico, USA, April 29 - May 4, 2025},
  pages        = {6859--6877},
  publisher    = {Association for Computational Linguistics},
  year         = {2025},
  url          = {https://doi.org/10.18653/v1/2025.naacl-long.350},
  doi          = {10.18653/V1/2025.NAACL-LONG.350},
  bibsource    = {dblp computer science bibliography, https://dblp.org}
}

@inproceedings{culture5,
  author       = {Jiahao Ying and
                  Wei Tang and
                  Yiran Zhao and
                  Yixin Cao and
                  Yu Rong and
                  Wenxuan Zhang},
  editor       = {Wanxiang Che and
                  Joyce Nabende and
                  Ekaterina Shutova and
                  Mohammad Taher Pilehvar},
  title        = {Disentangling Language and Culture for Evaluating Multilingual Large
                  Language Models},
  booktitle    = {Proceedings of the 63rd Annual Meeting of the Association for Computational
                  Linguistics (Volume 1: Long Papers), {ACL} 2025, Vienna, Austria,
                  July 27 - August 1, 2025},
  pages        = {22230--22251},
  publisher    = {Association for Computational Linguistics},
  year         = {2025},
  url          = {https://aclanthology.org/2025.acl-long.1082/},
  bibsource    = {dblp computer science bibliography, https://dblp.org}
}

@article{m-understand3,
  author       = {Weixiang Zhao and
                  Jiahe Guo and
                  Yang Deng and
                  Tongtong Wu and
                  Wenxuan Zhang and
                  Yulin Hu and
                  Xingyu Sui and
                  Yanyan Zhao and
                  Wanxiang Che and
                  Bing Qin and
                  Tat{-}Seng Chua and
                  Ting Liu},
  title        = {When Less Language is More: Language-Reasoning Disentanglement Makes
                  LLMs Better Multilingual Reasoners},
  journal      = {CoRR},
  volume       = {abs/2505.15257},
  year         = {2025},
  url          = {https://doi.org/10.48550/arXiv.2505.15257},
  doi          = {10.48550/ARXIV.2505.15257},
  eprinttype   = {arXiv},
  eprint       = {2505.15257},
  bibsource    = {dblp computer science bibliography, https://dblp.org}
}

@inproceedings{m-understand2,
  author       = {Yiran Zhao and
                  Wenxuan Zhang and
                  Guizhen Chen and
                  Kenji Kawaguchi and
                  Lidong Bing},
  editor       = {Amir Globersons and
                  Lester Mackey and
                  Danielle Belgrave and
                  Angela Fan and
                  Ulrich Paquet and
                  Jakub M. Tomczak and
                  Cheng Zhang},
  title        = {How do Large Language Models Handle Multilingualism?},
  booktitle    = {Advances in Neural Information Processing Systems 38: Annual Conference
                  on Neural Information Processing Systems 2024, NeurIPS 2024, Vancouver,
                  BC, Canada, December 10 - 15, 2024},
  year         = {2024},
  url          = {http://papers.nips.cc/paper\_files/paper/2024/hash/1bd359b32ab8b2a6bbafa1ed2856cf40-Abstract-Conference.html},
  bibsource    = {dblp computer science bibliography, https://dblp.org}
}

@article{m-understand1,
  author       = {Yuxin Chen and
                  Yiran Zhao and
                  Yang Zhang and
                  An Zhang and
                  Kenji Kawaguchi and
                  Shafiq Joty and
                  Junnan Li and
                  Tat{-}Seng Chua and
                  Michael Qizhe Shieh and
                  Wenxuan Zhang},
  title        = {The Emergence of Abstract Thought in Large Language Models Beyond
                  Any Language},
  journal      = {CoRR},
  volume       = {abs/2506.09890},
  year         = {2025},
  url          = {https://doi.org/10.48550/arXiv.2506.09890},
  doi          = {10.48550/ARXIV.2506.09890},
  eprinttype   = {arXiv},
  eprint       = {2506.09890},
  bibsource    = {dblp computer science bibliography, https://dblp.org}
}

@article{m-help,
  author       = {Changjiang Gao and
                  Zixian Huang and
                  Kaichen Yang and
                  Jiajun Chen and
                  Jixing Li and
                  Shujian Huang},
  title        = {ExpLang: Improved Exploration and Exploitation in {LLM} Reasoning
                  with On-Policy Thinking Language Selection},
  journal      = {CoRR},
  volume       = {abs/2602.21887},
  year         = {2026},
  url          = {https://doi.org/10.48550/arXiv.2602.21887},
  doi          = {10.48550/ARXIV.2602.21887},
  eprinttype   = {arXiv},
  eprint       = {2602.21887},
  bibsource    = {dblp computer science bibliography, https://dblp.org}
}

@article{babel,
  author       = {Yiran Zhao and
                  Chaoqun Liu and
                  Yue Deng and
                  Jiahao Ying and
                  Mahani Aljunied and
                  Zhaodonghui Li and
                  Lidong Bing and
                  Hou Pong Chan and
                  Yu Rong and
                  Deli Zhao and
                  Wenxuan Zhang},
  title        = {Babel: Open Multilingual Large Language Models Serving Over 90{\%}
                  of Global Speakers},
  journal      = {CoRR},
  volume       = {abs/2503.00865},
  year         = {2025},
  url          = {https://doi.org/10.48550/arXiv.2503.00865},
  doi          = {10.48550/ARXIV.2503.00865},
  eprinttype   = {arXiv},
  eprint       = {2503.00865},
  bibsource    = {dblp computer science bibliography, https://dblp.org}
}

@inproceedings{seallm,
  author       = {Wenxuan Zhang and
                  Hou Pong Chan and
                  Yiran Zhao and
                  Mahani Aljunied and
                  Jianyu Wang and
                  Chaoqun Liu and
                  Yue Deng and
                  Zhiqiang Hu and
                  Weiwen Xu and
                  Yew Ken Chia and
                  Xin Li and
                  Lidong Bing},
  editor       = {Nouha Dziri and
                  Sean (Xiang) Ren and
                  Shizhe Diao},
  title        = {SeaLLMs 3: Open Foundation and Chat Multilingual Large Language Models
                  for Southeast Asian Languages},
  booktitle    = {Proceedings of the 2025 Conference of the Nations of the Americas
                  Chapter of the Association for Computational Linguistics: Human Language
                  Technologies, {NAACL} 2025 - System Demonstrations, Albuquerque, New
                  Mexico, USA, April 29 - May 4, 2025},
  pages        = {96--105},
  publisher    = {Association for Computational Linguistics},
  year         = {2025},
  url          = {https://doi.org/10.18653/v1/2025.naacl-demo.10},
  doi          = {10.18653/V1/2025.NAACL-DEMO.10},
  bibsource    = {dblp computer science bibliography, https://dblp.org}
}

@inproceedings{mbert,
  author       = {Telmo Pires and
                  Eva Schlinger and
                  Dan Garrette},
  editor       = {Anna Korhonen and
                  David R. Traum and
                  Llu{\'{\i}}s M{\`{a}}rquez},
  title        = {How Multilingual is Multilingual BERT?},
  booktitle    = {Proceedings of the 57th Conference of the Association for Computational
                  Linguistics, {ACL} 2019, Florence, Italy, July 28- August 2, 2019,
                  Volume 1: Long Papers},
  pages        = {4996--5001},
  publisher    = {Association for Computational Linguistics},
  year         = {2019},
  url          = {https://doi.org/10.18653/v1/p19-1493},
  doi          = {10.18653/V1/P19-1493},
  bibsource    = {dblp computer science bibliography, https://dblp.org}
}

@article{wvs,
  author       = {Christian Haerpfer and
                  Ronald Inglehart and
                  Alejandro Moreno and
                  Christian Welzel and
                  Kseniya Kizilova and
                  Jaime Diez-Medrano and
                  Marta Lagos and
                  Pippa Norris and
                  Eduard Ponarin and
                  Bjorn Puranen},
  title        = {World values survey: Round seven-country-pooled datafile version 5.0},
  journal      = {Madrid, Spain \& Vienna, Austria: JD Systems Institute \& WVSA Secretariat},
  volume={12},
  number={10},
  pages={8},
  year={2022}
}

@inproceedings{blend,
  author       = {Junho Myung and
                  Nayeon Lee and
                  Yi Zhou and
                  Jiho Jin and
                  Rifki Afina Putri and
                  Dimosthenis Antypas and
                  Hsuvas Borkakoty and
                  Eunsu Kim and
                  Carla P{\'{e}}rez{-}Almendros and
                  Abinew Ali Ayele and
                  V{\'{\i}}ctor Guti{\'{e}}rrez{-}Basulto and
                  Yazm{\'{\i}}n Ib{\'{a}}{\~{n}}ez{-}Garc{\'{\i}}a and
                  Hwaran Lee and
                  Shamsuddeen Hassan Muhammad and
                  Ki{-}Woong Park and
                  Anar Rzayev and
                  Nina White and
                  Seid Muhie Yimam and
                  Mohammad Taher Pilehvar and
                  Nedjma Ousidhoum and
                  Jos{\'{e}} Camacho{-}Collados and
                  Alice Oh},
  editor       = {Amir Globersons and
                  Lester Mackey and
                  Danielle Belgrave and
                  Angela Fan and
                  Ulrich Paquet and
                  Jakub M. Tomczak and
                  Cheng Zhang},
  title        = {BLEnD: {A} Benchmark for LLMs on Everyday Knowledge in Diverse Cultures
                  and Languages},
  booktitle    = {Advances in Neural Information Processing Systems 38: Annual Conference
                  on Neural Information Processing Systems 2024, NeurIPS 2024, Vancouver,
                  BC, Canada, December 10 - 15, 2024},
  year         = {2024},
  url          = {http://papers.nips.cc/paper\_files/paper/2024/hash/8eb88844dafefa92a26aaec9f3acad93-Abstract-Datasets\_and\_Benchmarks\_Track.html},
  bibsource    = {dblp computer science bibliography, https://dblp.org}
}

@article{NIPS,
  author       = {Liwei Jiang and
                  Yuanjun Chai and
                  Margaret Li and
                  Mickel Liu and
                  Raymond Fok and
                  Nouha Dziri and
                  Yulia Tsvetkov and
                  Maarten Sap and
                  Alon Albalak and
                  Yejin Choi},
  title        = {Artificial Hivemind: The Open-Ended Homogeneity of Language Models
                  (and Beyond)},
  journal      = {CoRR},
  volume       = {abs/2510.22954},
  year         = {2025},
  url          = {https://doi.org/10.48550/arXiv.2510.22954},
  doi          = {10.48550/ARXIV.2510.22954},
  eprinttype    = {arXiv},
  eprint       = {2510.22954},
  bibsource    = {dblp computer science bibliography, https://dblp.org}
}

@inproceedings{culture1,
  author       = {Badr AlKhamissi and
                  Muhammad N. ElNokrashy and
                  Mai Alkhamissi and
                  Mona T. Diab},
  editor       = {Lun{-}Wei Ku and
                  Andre Martins and
                  Vivek Srikumar},
  title        = {Investigating Cultural Alignment of Large Language Models},
  booktitle    = {Proceedings of the 62nd Annual Meeting of the Association for Computational
                  Linguistics (Volume 1: Long Papers), {ACL} 2024, Bangkok, Thailand,
                  August 11-16, 2024},
  pages        = {12404--12422},
  publisher    = {Association for Computational Linguistics},
  year         = {2024},
  url          = {https://doi.org/10.18653/v1/2024.acl-long.671},
  doi          = {10.18653/V1/2024.ACL-LONG.671},
  bibsource    = {dblp computer science bibliography, https://dblp.org}
}

@inproceedings{culture2,
  author       = {Wenxuan Wang and
                  Wenxiang Jiao and
                  Jingyuan Huang and
                  Ruyi Dai and
                  Jen{-}tse Huang and
                  Zhaopeng Tu and
                  Michael R. Lyu},
  editor       = {Lun{-}Wei Ku and
                  Andre Martins and
                  Vivek Srikumar},
  title        = {Not All Countries Celebrate Thanksgiving: On the Cultural Dominance
                  in Large Language Models},
  booktitle    = {Proceedings of the 62nd Annual Meeting of the Association for Computational
                  Linguistics (Volume 1: Long Papers), {ACL} 2024, Bangkok, Thailand,
                  August 11-16, 2024},
  pages        = {6349--6384},
  publisher    = {Association for Computational Linguistics},
  year         = {2024},
  url          = {https://doi.org/10.18653/v1/2024.acl-long.345},
  doi          = {10.18653/V1/2024.ACL-LONG.345},
  bibsource    = {dblp computer science bibliography, https://dblp.org}
}

@inproceedings{plural,
  author       = {Taylor Sorensen and
                  Jared Moore and
                  Jillian Fisher and
                  Mitchell L. Gordon and
                  Niloofar Mireshghallah and
                  Christopher Michael Rytting and
                  Andre Ye and
                  Liwei Jiang and
                  Ximing Lu and
                  Nouha Dziri and
                  Tim Althoff and
                  Yejin Choi},
  title        = {Position: {A} Roadmap to Pluralistic Alignment},
  booktitle    = {Forty-first International Conference on Machine Learning, {ICML} 2024,
                  Vienna, Austria, July 21-27, 2024},
  publisher    = {OpenReview.net},
  year         = {2024},
  url          = {https://openreview.net/forum?id=gQpBnRHwxM},
  bibsource    = {dblp computer science bibliography, https://dblp.org}
}

@inproceedings{plural2,
  author       = {Vincent Conitzer and
                  Rachel Freedman and
                  Jobst Heitzig and
                  Wesley H. Holliday and
                  Bob M. Jacobs and
                  Nathan Lambert and
                  Milan Moss{\'{e}} and
                  Eric Pacuit and
                  Stuart Russell and
                  Hailey Schoelkopf and
                  Emanuel Tewolde and
                  William S. Zwicker},
  title        = {Position: Social Choice Should Guide {AI} Alignment in Dealing with
                  Diverse Human Feedback},
  booktitle    = {Forty-first International Conference on Machine Learning, {ICML} 2024,
                  Vienna, Austria, July 21-27, 2024},
  year         = {2024},
  url          = {https://openreview.net/forum?id=w1d9DOGymR},
  bibsource    = {dblp computer science bibliography, https://dblp.org}
}

@article{traditional1,
  author       = {Yanzhu Guo and
                  Guokan Shang and
                  Chlo{\'{e}} Clavel},
  title        = {Benchmarking Linguistic Diversity of Large Language Models},
  journal      = {Trans. Assoc. Comput. Linguistics},
  volume       = {13},
  pages        = {1507--1526},
  year         = {2025},
  url          = {https://doi.org/10.1162/tacl.a.47},
  doi          = {10.1162/TACL.A.47},
  bibsource    = {dblp computer science bibliography, https://dblp.org}
}

@article{traditional2,
  author       = {Antoine Bellemare P{\'{e}}pin and
                  Fran{\c{c}}ois Lespinasse and
                  Philipp Th{\"{o}}lke and
                  Yann Harel and
                  Kory W. Mathewson and
                  Jay A. Olson and
                  Yoshua Bengio and
                  Karim Jerbi},
  title        = {Divergent Creativity in Humans and Large Language Models},
  journal      = {CoRR},
  volume       = {abs/2405.13012},
  year         = {2024},
  url          = {https://doi.org/10.48550/arXiv.2405.13012},
  doi          = {10.48550/ARXIV.2405.13012},
  eprinttype    = {arXiv},
  eprint       = {2405.13012},
  bibsource    = {dblp computer science bibliography, https://dblp.org}
}

@article{chat1,
  author       = {Yiming Zhang and
                  Harshita Diddee and
                  Susan Holm and
                  Hanchen Liu and
                  Xinyue Liu and
                  Vinay Samuel and
                  Barry Wang and
                  Daphne Ippolito},
  title        = {NoveltyBench: Evaluating Language Models for Humanlike Diversity},
  journal      = {CoRR},
  volume       = {abs/2504.05228},
  year         = {2025},
  url          = {https://doi.org/10.48550/arXiv.2504.05228},
  doi          = {10.48550/ARXIV.2504.05228},
  eprinttype    = {arXiv},
  eprint       = {2504.05228},
  bibsource    = {dblp computer science bibliography, https://dblp.org}
}

@article{chat2,
  author       = {Arash Lagzian and
                  Srinivas Anumasa and
                  Dianbo Liu},
  title        = {Multi-Novelty: Improve the Diversity and Novelty of Contents Generated
                  by Large Language Models via inference-time Multi-Views Brainstorming},
  journal      = {CoRR},
  volume       = {abs/2502.12700},
  year         = {2025},
  url          = {https://doi.org/10.48550/arXiv.2502.12700},
  doi          = {10.48550/ARXIV.2502.12700},
  eprinttype    = {arXiv},
  eprint       = {2502.12700},
  bibsource    = {dblp computer science bibliography, https://dblp.org}
}

@inproceedings{math1,
  author       = {Junyi Ye and
                  Jingyi Gu and
                  Xinyun Zhao and
                  Wenpeng Yin and
                  Grace Guiling Wang},
  editor       = {Toby Walsh and
                  Julie Shah and
                  Zico Kolter},
  title        = {Assessing the Creativity of LLMs in Proposing Novel Solutions to Mathematical
                  Problems},
  booktitle    = {AAAI-25, Sponsored by the Association for the Advancement of Artificial
                  Intelligence, February 25 - March 4, 2025, Philadelphia, PA, {USA}},
  pages        = {25687--25696},
  publisher    = {{AAAI} Press},
  year         = {2025},
  url          = {https://doi.org/10.1609/aaai.v39i24.34760},
  doi          = {10.1609/AAAI.V39I24.34760},
  bibsource    = {dblp computer science bibliography, https://dblp.org}
}

@inproceedings{math2,
  author       = {Yufei Tian and
                  Abhilasha Ravichander and
                  Lianhui Qin and
                  Ronan Le Bras and
                  Raja Marjieh and
                  Nanyun Peng and
                  Yejin Choi and
                  Thomas L. Griffiths and
                  Faeze Brahman},
  editor       = {Kevin Duh and
                  Helena G{\'{o}}mez{-}Adorno and
                  Steven Bethard},
  title        = {MacGyver: Are Large Language Models Creative Problem Solvers?},
  booktitle    = {Proceedings of the 2024 Conference of the North American Chapter of
                  the Association for Computational Linguistics: Human Language Technologies
                  (Volume 1: Long Papers), {NAACL} 2024, Mexico City, Mexico, June 16-21,
                  2024},
  pages        = {5303--5324},
  publisher    = {Association for Computational Linguistics},
  year         = {2024},
  url          = {https://doi.org/10.18653/v1/2024.naacl-long.297},
  doi          = {10.18653/V1/2024.NAACL-LONG.297},
  bibsource    = {dblp computer science bibliography, https://dblp.org}
}

@article{math3,
  author       = {Xiaoyang Chen and
                  Xinan Dai and
                  Yu Du and
                  Qian Feng and
                  Naixu Guo and
                  Tingshuo Gu and
                  Yuting Gao and
                  Yingyi Gao and
                  Xudong Han and
                  Xiang Jiang and
                  Yilin Jin and
                  Hongyi Lin and
                  Shisheng Lin and
                  Xiangnan Li and
                  Yuante Li and
                  Yixing Li and
                  Zhentao Lai and
                  Zilu Ma and
                  Yingrong Peng and
                  Jiacheng Qian and
                  Hao{-}Yu Sun and
                  Jianbo Sun and
                  Zirui Wang and
                  Siwei Wu and
                  Zian Wang and
                  Bin Xu and
                  Jianghao Xu and
                  Yiyang Yu and
                  Zichuan Yang and
                  Hongji Zha and
                  Ruichong Zhang},
  title        = {DeepMath-Creative: {A} Benchmark for Evaluating Mathematical Creativity
                  of Large Language Models},
  journal      = {CoRR},
  volume       = {abs/2505.08744},
  year         = {2025},
  url          = {https://doi.org/10.48550/arXiv.2505.08744},
  doi          = {10.48550/ARXIV.2505.08744},
  eprinttype    = {arXiv},
  eprint       = {2505.08744},
  bibsource    = {dblp computer science bibliography, https://dblp.org}
}

@article{math4,
  author       = {Simeng Han and
                  Stephen Xia and
                  Grant Zhang and
                  Howard Dai and
                  Chen Liu and
                  Lichang Chen and
                  Hoang Huy Nguyen and
                  Hongyuan Mei and
                  Jiayuan Mao and
                  R. Thomas McCoy},
  title        = {Creativity or Brute Force? Using Brainteasers as a Window into the
                  Problem-Solving Abilities of Large Language Models},
  journal      = {CoRR},
  volume       = {abs/2505.10844},
  year         = {2025},
  url          = {https://doi.org/10.48550/arXiv.2505.10844},
  doi          = {10.48550/ARXIV.2505.10844},
  eprinttype    = {arXiv},
  eprint       = {2505.10844},
  bibsource    = {dblp computer science bibliography, https://dblp.org}
}

@inproceedings{idea1,
  author       = {Sikun Guo and
                  Amir Hassan Shariatmadari and
                  Guangzhi Xiong and
                  Albert Huang and
                  Myles Kim and
                  Corey M. Williams and
                  Stefan Bekiranov and
                  Aidong Zhang},
  editor       = {Luiza Antonie and
                  Jian Pei and
                  Xiaohui Yu and
                  Flavio Chierichetti and
                  Hady W. Lauw and
                  Yizhou Sun and
                  Srinivasan Parthasarathy},
  title        = {IdeaBench: Benchmarking Large Language Models for Research Idea Generation},
  booktitle    = {Proceedings of the 31st {ACM} {SIGKDD} Conference on Knowledge Discovery
                  and Data Mining, V.2, {KDD} 2025, Toronto ON, Canada, August 3-7,
                  2025},
  pages        = {5888--5899},
  publisher    = {{ACM}},
  year         = {2025},
  url          = {https://doi.org/10.1145/3711896.3737419},
  doi          = {10.1145/3711896.3737419},
  bibsource    = {dblp computer science bibliography, https://dblp.org}
}

@article{idea2,
  author       = {Kai Ruan and
                  Xuan Wang and
                  Jixiang Hong and
                  Peng Wang and
                 Yang Liu and
                  Hao Sun},
  title        = {LiveIdeaBench: Evaluating LLMs' Divergent Thinking for Scientific Idea Generation with Minimal Context},
  journal      = {CoRR},
  volume       = {abs/2412.17596},
  year         = {2024},
  url          = {https://doi.org/10.48550/arXiv.2412.17596},
  doi          = {10.48550/ARXIV.2412.17596},
  eprinttype    = {arXiv},
  eprint       = {2412.17596},
}

@inproceedings{influence1,
  author       = {Max Peeperkorn and
                  Tom Kouwenhoven and
                  Dan Brown and
                  Anna Jordanous},
  editor       = {Kazjon Grace and
                  Maria Teresa Llano and
                  Pedro Martins and
                  Maria M. Hedblom},
  title        = {Is Temperature the Creativity Parameter of Large Language Models?},
  booktitle    = {Proceedings of the 15th International Conference on Computational
                  Creativity, {ICCC} 2024, J{\"{o}}nk{\"{o}}ping, Sweden,
                  June 17-21, 2024},
  pages        = {226--235},
  publisher    = {Association for Computational Creativity {(ACC)}},
  year         = {2024},
  url          = {https://computationalcreativity.net/iccc24/papers/ICCC24\_paper\_70.pdf},
  bibsource    = {dblp computer science bibliography, https://dblp.org}
}

@inproceedings{influence3,
  author       = {Ziniu Li and
                  Congliang Chen and
                  Tian Xu and
                  Zeyu Qin and
                  Jiancong Xiao and
                  Zhi{-}Quan Luo and
                  Ruoyu Sun},
  title        = {Preserving Diversity in Supervised Fine-Tuning of Large Language Models},
  booktitle    = {The Thirteenth International Conference on Learning Representations,
                  {ICLR} 2025, Singapore, April 24-28, 2025},
  publisher    = {OpenReview.net},
  year         = {2025},
  url          = {https://openreview.net/forum?id=NQEe7B7bSw},
  bibsource    = {dblp computer science bibliography, https://dblp.org}
}

@inproceedings{influence4,
  author       = {Haoran Sun and
                  Yekun Chai and
                  Shuohuan Wang and
                  Yu Sun and
                  Hua Wu and
                  Haifeng Wang},
  editor       = {Wanxiang Che and
                  Joyce Nabende and
                  Ekaterina Shutova and
                  Mohammad Taher Pilehvar},
  title        = {Curiosity-Driven Reinforcement Learning from Human Feedback},
  booktitle    = {Proceedings of the 63rd Annual Meeting of the Association for Computational
                  Linguistics (Volume 1: Long Papers), {ACL} 2025, Vienna, Austria,
                  July 27 - August 1, 2025},
  pages        = {23517--23534},
  publisher    = {Association for Computational Linguistics},
  year         = {2025},
  url          = {https://aclanthology.org/2025.acl-long.1146/},
  bibsource    = {dblp computer science bibliography, https://dblp.org}
}

@article{theory1,
  title={Over-reliance on English hinders cognitive science},
  author={Damián E. Blasi and
          Joseph Henrich and
          Evangelia Adamou and
          David Kemmerer and
         Asifa Majid},
  journal={Trends in Cognitive Sciences},
  volume={26},
  number={12},
  pages={1153--1170},
  year={2022},
}

@article{theory2,
  title={The effects of multilingual and multicultural practices on divergent thinking. Implications for plurilingual creativity paradigm},
  author={Anatoliy V. Kharkhurin and
          Valeriya Koncha and
          Morteza Charkhabi},
  journal={Bilingualism: Language and cognition},
  volume={26},
  number={3},
  pages={592--609},
  year={2023},
}

@book{theory3,
  title     = {Language, Thought, and Reality: Selected Writings of Benjamin Lee Whorf},
  author    = {Whorf, Benjamin Lee},
  year      = {2012},
  publisher = {MIT Press}
}

@inproceedings{eval1,
  author       = {Vishakh Padmakumar and
                  He He},
  title        = {Does Writing with Language Models Reduce Content Diversity?},
  booktitle    = {The Twelfth International Conference on Learning Representations,
                  {ICLR} 2024, Vienna, Austria, May 7-11, 2024},
  publisher    = {OpenReview.net},
  year         = {2024},
  url          = {https://openreview.net/forum?id=Feiz5HtCD0},
  bibsource    = {dblp computer science bibliography, https://dblp.org}
}

@inproceedings{eval2,
  author       = {Yanzhu Guo and
                  Guokan Shang and
                  Michalis Vazirgiannis and
                  Chlo{\'{e}} Clavel},
  editor       = {Kevin Duh and
                  Helena G{\'{o}}mez{-}Adorno and
                  Steven Bethard},
  title        = {The Curious Decline of Linguistic Diversity: Training Language Models
                  on Synthetic Text},
  booktitle    = {Findings of the Association for Computational Linguistics: {NAACL}
                  2024, Mexico City, Mexico, June 16-21, 2024},
  pages        = {3589--3604},
  publisher    = {Association for Computational Linguistics},
  year         = {2024},
  url          = {https://doi.org/10.18653/v1/2024.findings-naacl.228},
  doi          = {10.18653/V1/2024.FINDINGS-NAACL.228},
  bibsource    = {dblp computer science bibliography, https://dblp.org}
}

@inproceedings{eval5,
  author       = {Nils Reimers and
                  Iryna Gurevych},
  editor       = {Kentaro Inui and
                  Jing Jiang and
                  Vincent Ng and
                  Xiaojun Wan},
  title        = {Sentence-BERT: Sentence Embeddings using Siamese BERT-Networks},
  booktitle    = {Proceedings of the 2019 Conference on Empirical Methods in Natural
                  Language Processing and the 9th International Joint Conference on
                  Natural Language Processing, {EMNLP-IJCNLP} 2019, Hong Kong, China,
                  November 3-7, 2019},
  pages        = {3980--3990},
  publisher    = {Association for Computational Linguistics},
  year         = {2019},
  url          = {https://doi.org/10.18653/v1/D19-1410},
  doi          = {10.18653/V1/D19-1410},
  bibsource    = {dblp computer science bibliography, https://dblp.org}
}

@inproceedings{eval6,
  author       = {Yaoming Zhu and
                  Sidi Lu and
                  Lei Zheng and
                  Jiaxian Guo and
                  Weinan Zhang and
                  Jun Wang and
                  Yong Yu},
  editor       = {Kevyn Collins{-}Thompson and
                  Qiaozhu Mei and
                  Brian D. Davison and
                  Yiqun Liu and
                  Emine Yilmaz},
  title        = {Texygen: {A} Benchmarking Platform for Text Generation Models},
  booktitle    = {The 41st International {ACM} {SIGIR} Conference on Research {\&}
                  Development in Information Retrieval, {SIGIR} 2018, Ann Arbor, MI,
                  USA, July 08-12, 2018},
  pages        = {1097--1100},
  publisher    = {{ACM}},
  year         = {2018},
  url          = {https://doi.org/10.1145/3209978.3210080},
  doi          = {10.1145/3209978.3210080},
  bibsource    = {dblp computer science bibliography, https://dblp.org}
}

@article{eval7,
  author       = {Weixin Liang and
                  Yaohui Zhang and
                  Zhengxuan Wu and
                  Haley Lepp and
                  Wenlong Ji and
                  Xuandong Zhao and
                  Hancheng Cao and
                  Sheng Liu and
                  Siyu He and
                  Zhi Huang and
                  Diyi Yang and
                  Christopher Potts and
                  Christopher D. Manning and
                  James Y. Zou},
  title        = {Mapping the Increasing Use of LLMs in Scientific Papers},
  journal      = {CoRR},
  volume       = {abs/2404.01268},
  year         = {2024},
  url          = {https://doi.org/10.48550/arXiv.2404.01268},
  doi          = {10.48550/ARXIV.2404.01268},
  eprinttype    = {arXiv},
  eprint       = {2404.01268},
  bibsource    = {dblp computer science bibliography, https://dblp.org}
}

@article{eval8,
  author       = {Jiaming Luo and
                  Colin Cherry and
                  George F. Foster},
  title        = {To Diverge or Not to Diverge: {A} Morphosyntactic Perspective on Machine
                  Translation vs Human Translation},
  journal      = {Trans. Assoc. Comput. Linguistics},
  volume       = {12},
  pages        = {355--371},
  year         = {2024},
  url          = {https://doi.org/10.1162/tacl\_a\_00645},
  doi          = {10.1162/TACL\_A\_00645},
  bibsource    = {dblp computer science bibliography, https://dblp.org}
}

@inproceedings{eval9,
  author       = {Salvatore Giorgi and
                  Tingting Liu and
                  Ankit Aich and
                  Kelsey Isman and
                  Garrick Sherman and
                  Zachary Fried and
                  Jo{\~{a}}o Sedoc and
                  Lyle H. Ungar and
                  Brenda Curtis},
  editor       = {Yaser Al{-}Onaizan and
                  Mohit Bansal and
                  Yun{-}Nung Chen},
  title        = {Modeling Human Subjectivity in LLMs Using Explicit and Implicit Human
                  Factors in Personas},
  booktitle    = {Findings of the Association for Computational Linguistics: {EMNLP}
                  2024, Miami, Florida, USA, November 12-16, 2024},
  pages        = {7174--7188},
  publisher    = {Association for Computational Linguistics},
  year         = {2024},
  url          = {https://doi.org/10.18653/v1/2024.findings-emnlp.420},
  doi          = {10.18653/V1/2024.FINDINGS-EMNLP.420},
  bibsource    = {dblp computer science bibliography, https://dblp.org}
}

@inproceedings{multi1,
  author       = {Tian Liang and
                  Zhiwei He and
                  Wenxiang Jiao and
                  Xing Wang and
                  Yan Wang and
                  Rui Wang and
                  Yujiu Yang and
                  Shuming Shi and
                  Zhaopeng Tu},
  editor       = {Yaser Al{-}Onaizan and
                  Mohit Bansal and
                  Yun{-}Nung Chen},
  title        = {Encouraging Divergent Thinking in Large Language Models through Multi-Agent
                  Debate},
  booktitle    = {Proceedings of the 2024 Conference on Empirical Methods in Natural
                  Language Processing, {EMNLP} 2024, Miami, FL, USA, November 12-16,
                  2024},
  pages        = {17889--17904},
  publisher    = {Association for Computational Linguistics},
  year         = {2024},
  url          = {https://doi.org/10.18653/v1/2024.emnlp-main.992},
  doi          = {10.18653/V1/2024.EMNLP-MAIN.992},
  bibsource    = {dblp computer science bibliography, https://dblp.org}
}

@inproceedings{temp1,
  author       = {Guy Tevet and
                  Jonathan Berant},
  editor       = {Paola Merlo and
                  J{\"{o}}rg Tiedemann and
                  Reut Tsarfaty},
  title        = {Evaluating the Evaluation of Diversity in Natural Language Generation},
  booktitle    = {Proceedings of the 16th Conference of the European Chapter of the
                  Association for Computational Linguistics: Main Volume, {EACL} 2021,
                  Online, April 19 - 23, 2021},
  pages        = {326--346},
  publisher    = {Association for Computational Linguistics},
  year         = {2021},
  url          = {https://doi.org/10.18653/v1/2021.eacl-main.25},
  doi          = {10.18653/V1/2021.EACL-MAIN.25},
  bibsource    = {dblp computer science bibliography, https://dblp.org}
}

@article{multi2,
  author       = {Michal Shur{-}Ofry and
                  Bar Horowitz{-}Amsalem and
                  Adir Rahamim and
                  Yonatan Belinkov},
  title        = {Growing a Tail: Increasing Output Diversity in Large Language Models},
  journal      = {CoRR},
  volume       = {abs/2411.02989},
  year         = {2024},
  url          = {https://doi.org/10.48550/arXiv.2411.02989},
  doi          = {10.48550/ARXIV.2411.02989},
  eprinttype    = {arXiv},
  eprint       = {2411.02989},
  bibsource    = {dblp computer science bibliography, https://dblp.org}
}

@inproceedings{multi3,
  author       = {Qihan Wang and
                  Shidong Pan and
                  Tal Linzen and
                  Emily Black},
  editor       = {Christos Christodoulopoulos and
                  Tanmoy Chakraborty and
                  Carolyn Rose and
                  Violet Peng},
  title        = {Multilingual Prompting for Improving {LLM} Generation Diversity},
  booktitle    = {Proceedings of the 2025 Conference on Empirical Methods in Natural
                  Language Processing, {EMNLP} 2025, Suzhou, China, November 4-9, 2025},
  pages        = {6367--6389},
  publisher    = {Association for Computational Linguistics},
  year         = {2025},
  url          = {https://doi.org/10.18653/v1/2025.emnlp-main.324},
  doi          = {10.18653/V1/2025.EMNLP-MAIN.324},
  bibsource    = {dblp computer science bibliography, https://dblp.org}
}

@inproceedings{tts1,
  author       = {Niklas Muennighoff and
                  Zitong Yang and
                  Weijia Shi and
                  Xiang Lisa Li and
                  Li Fei{-}Fei and
                  Hannaneh Hajishirzi and
                  Luke Zettlemoyer and
                  Percy Liang and
                  Emmanuel J. Cand{\`{e}}s and
                  Tatsunori Hashimoto},
  editor       = {Christos Christodoulopoulos and
                  Tanmoy Chakraborty and
                  Carolyn Rose and
                  Violet Peng},
  title        = {s1: Simple test-time scaling},
  booktitle    = {Proceedings of the 2025 Conference on Empirical Methods in Natural
                  Language Processing, {EMNLP} 2025, Suzhou, China, November 4-9, 2025},
  pages        = {20275--20321},
  publisher    = {Association for Computational Linguistics},
  year         = {2025},
  url          = {https://doi.org/10.18653/v1/2025.emnlp-main.1025},
  doi          = {10.18653/V1/2025.EMNLP-MAIN.1025},
  bibsource    = {dblp computer science bibliography, https://dblp.org}
}

@inproceedings{tts2,
  author       = {Zhiyuan Zeng and
                  Qinyuan Cheng and
                  Zhangyue Yin and
                  Yunhua Zhou and
                  Xipeng Qiu},
  editor       = {Wanxiang Che and
                  Joyce Nabende and
                  Ekaterina Shutova and
                  Mohammad Taher Pilehvar},
  title        = {Revisiting the Test-Time Scaling of o1-like Models: Do they Truly
                  Possess Test-Time Scaling Capabilities?},
  booktitle    = {Proceedings of the 63rd Annual Meeting of the Association for Computational
                  Linguistics (Volume 1: Long Papers), {ACL} 2025, Vienna, Austria,
                  July 27 - August 1, 2025},
  pages        = {4651--4665},
  publisher    = {Association for Computational Linguistics},
  year         = {2025},
  url          = {https://aclanthology.org/2025.acl-long.232/},
  bibsource    = {dblp computer science bibliography, https://dblp.org}
}

@article{tts3,
  author       = {DeepSeek{-}AI and
                  Daya Guo and
                  Dejian Yang and
                  Haowei Zhang and
                  Junxiao Song and
                  Ruoyu Zhang and
                  Runxin Xu and
                  Qihao Zhu and
                  Shirong Ma and
                  Peiyi Wang and
                  Xiao Bi and
                  Xiaokang Zhang and
                  Xingkai Yu and
                  Yu Wu and
                  Z. F. Wu and
                  Zhibin Gou and
                  Zhihong Shao and
                  Zhuoshu Li and
                  Ziyi Gao and
                  Aixin Liu and
                  Bing Xue and
                  Bingxuan Wang and
                  Bochao Wu and
                  Bei Feng and
                  Chengda Lu and
                  Chenggang Zhao and
                  Chengqi Deng and
                  Chenyu Zhang and
                  Chong Ruan and
                  Damai Dai and
                  Deli Chen and
                  Dongjie Ji and
                  Erhang Li and
                  Fangyun Lin and
                  Fucong Dai and
                  Fuli Luo and
                  Guangbo Hao and
                  Guanting Chen and
                  Guowei Li and
                  H. Zhang and
                  Han Bao and
                  Hanwei Xu and
                  Haocheng Wang and
                  Honghui Ding and
                  Huajian Xin and
                  Huazuo Gao and
                  Hui Qu and
                  Hui Li and
                  Jianzhong Guo and
                  Jiashi Li and
                  Jiawei Wang and
                  Jingchang Chen and
                  Jingyang Yuan and
                  Junjie Qiu and
                  Junlong Li and
                  J. L. Cai and
                  Jiaqi Ni and
                  Jian Liang and
                  Jin Chen and
                  Kai Dong and
                  Kai Hu and
                  Kaige Gao and
                  Kang Guan and
                  Kexin Huang and
                  Kuai Yu and
                  Lean Wang and
                  Lecong Zhang and
                  Liang Zhao and
                  Litong Wang and
                  Liyue Zhang and
                  Lei Xu and
                  Leyi Xia and
                  Mingchuan Zhang and
                  Minghua Zhang and
                  Minghui Tang and
                  Meng Li and
                  Miaojun Wang and
                  Mingming Li and
                  Ning Tian and
                  Panpan Huang and
                  Peng Zhang and
                  Qiancheng Wang and
                  Qinyu Chen and
                  Qiushi Du and
                  Ruiqi Ge and
                  Ruisong Zhang and
                  Ruizhe Pan and
                  Runji Wang and
                  R. J. Chen and
                  R. L. Jin and
                  Ruyi Chen and
                  Shanghao Lu and
                  Shangyan Zhou and
                  Shanhuang Chen and
                  Shengfeng Ye and
                  Shiyu Wang and
                  Shuiping Yu and
                  Shunfeng Zhou and
                  Shuting Pan and
                  S. S. Li},
  title        = {DeepSeek-R1: Incentivizing Reasoning Capability in LLMs via Reinforcement
                  Learning},
  journal      = {CoRR},
  volume       = {abs/2501.12948},
  year         = {2025},
  url          = {https://doi.org/10.48550/arXiv.2501.12948},
  doi          = {10.48550/ARXIV.2501.12948},
  eprinttype    = {arXiv},
  eprint       = {2501.12948},
  bibsource    = {dblp computer science bibliography, https://dblp.org}
}

@inproceedings{mtts1,
  author       = {Guijin Son and
                  Jiwoo Hong and
                  Hyunwoo Ko and
                  James Thorne},
  editor       = {Wanxiang Che and
                  Joyce Nabende and
                  Ekaterina Shutova and
                  Mohammad Taher Pilehvar},
  title        = {Linguistic Generalizability of Test-Time Scaling in Mathematical Reasoning},
  booktitle    = {Proceedings of the 63rd Annual Meeting of the Association for Computational
                  Linguistics (Volume 1: Long Papers), {ACL} 2025, Vienna, Austria,
                  July 27 - August 1, 2025},
  pages        = {14333--14368},
  publisher    = {Association for Computational Linguistics},
  year         = {2025},
  url          = {https://aclanthology.org/2025.acl-long.699/},
  bibsource    = {dblp computer science bibliography, https://dblp.org}
}

@article{mtts2,
  author       = {Zheng{-}Xin Yong and
                  Muhammad Farid Adilazuarda and
                  Jonibek Mansurov and
                  Ruochen Zhang and
                  Niklas Muennighoff and
                  Carsten Eickhoff and
                  Genta Indra Winata and
                  Julia Kreutzer and
                  Stephen H. Bach and
                  Alham Fikri Aji},
  title        = {Crosslingual Reasoning through Test-Time Scaling},
  journal      = {CoRR},
  volume       = {abs/2505.05408},
  year         = {2025},
  url          = {https://doi.org/10.48550/arXiv.2505.05408},
  doi          = {10.48550/ARXIV.2505.05408},
  eprinttype    = {arXiv},
  eprint       = {2505.05408},
  bibsource    = {dblp computer science bibliography, https://dblp.org}
}

@article{mtts3,
  author       = {Yiming Wang and
                  Pei Zhang and
                  Jialong Tang and
                  Haoran Wei and
                  Baosong Yang and
                  Rui Wang and
                  Chenshu Sun and
                  Feitong Sun and
                  Jiran Zhang and
                  Junxuan Wu and
                  Qiqian Cang and
                  Yichang Zhang and
                  Fei Huang and
                  Junyang Lin and
                  Fei Huang and
                  Jingren Zhou},
  title        = {PolyMath: Evaluating Mathematical Reasoning in Multilingual Contexts},
  journal      = {CoRR},
  volume       = {abs/2504.18428},
  year         = {2025},
  url          = {https://doi.org/10.48550/arXiv.2504.18428},
  doi          = {10.48550/ARXIV.2504.18428},
  eprinttype    = {arXiv},
  eprint       = {2504.18428},
  bibsource    = {dblp computer science bibliography, https://dblp.org}
}

@article{mtts4,
  author       = {Prasoon Bajpai and
                  Tanmoy Chakraborty},
  title        = {Multilingual Test-Time Scaling via Initial Thought Transfer},
  journal      = {CoRR},
  volume       = {abs/2505.15508},
  year         = {2025},
  url          = {https://doi.org/10.48550/arXiv.2505.15508},
  doi          = {10.48550/ARXIV.2505.15508},
  eprinttype    = {arXiv},
  eprint       = {2505.15508},
  bibsource    = {dblp computer science bibliography, https://dblp.org}
}

@inproceedings{mtts5,
  author       = {Jirui Qi and
                  Shan Chen and
                  Zidi Xiong and
                  Raquel Fern{\'{a}}ndez and
                  Danielle S. Bitterman and
                  Arianna Bisazza},
  editor       = {Christos Christodoulopoulos and
                  Tanmoy Chakraborty and
                  Carolyn Rose and
                  Violet Peng},
  title        = {When Models Reason in Your Language: Controlling Thinking Language
                  Comes at the Cost of Accuracy},
  booktitle    = {Findings of the Association for Computational Linguistics: {EMNLP}
                  2025, Suzhou, China, November 4-9, 2025},
  pages        = {20279--20296},
  publisher    = {Association for Computational Linguistics},
  year         = {2025},
  url          = {https://aclanthology.org/2025.findings-emnlp.1103/},
  bibsource    = {dblp computer science bibliography, https://dblp.org}
}

@article{mtts6,
  author       = {Zhi Rui Tam and
                  Cheng{-}Kuang Wu and
                  Yu Ying Chiu and
                  Chieh{-}Yen Lin and
                  Yun{-}Nung Chen and
                  Hung{-}yi Lee},
  title        = {Language Matters: How Do Multilingual Input and Reasoning Paths Affect
                  Large Reasoning Models?},
  journal      = {CoRR},
  volume       = {abs/2505.17407},
  year         = {2025},
  url          = {https://doi.org/10.48550/arXiv.2505.17407},
  doi          = {10.48550/ARXIV.2505.17407},
  eprinttype    = {arXiv},
  eprint       = {2505.17407},
  bibsource    = {dblp computer science bibliography, https://dblp.org}
}

@inproceedings{mtts7,
  author       = {Ammar Khairi and
                  Daniel D'souza and
                  Ye Shen and
                  Julia Kreutzer and
                  Sara Hooker},
  editor       = {Christos Christodoulopoulos and
                  Tanmoy Chakraborty and
                  Carolyn Rose and
                  Violet Peng},
  title        = {When Life Gives You Samples: The Benefits of Scaling up Inference
                  Compute for Multilingual LLMs},
  booktitle    = {Proceedings of the 2025 Conference on Empirical Methods in Natural
                  Language Processing, {EMNLP} 2025, Suzhou, China, November 4-9, 2025},
  pages        = {27559--27583},
  publisher    = {Association for Computational Linguistics},
  year         = {2025},
  url          = {https://doi.org/10.18653/v1/2025.emnlp-main.1402},
  doi          = {10.18653/V1/2025.EMNLP-MAIN.1402},
  bibsource    = {dblp computer science bibliography, https://dblp.org}
}

@inproceedings{mtts9,
  author       = {Yihao Li and
                  Jiayi Xin and
                  Miranda Muqing Miao and
                  Qi Long and
                  Lyle H. Ungar},
  editor       = {Christos Christodoulopoulos and
                  Tanmoy Chakraborty and
                  Carolyn Rose and
                  Violet Peng},
  title        = {The Impact of Language Mixing on Bilingual {LLM} Reasoning},
  booktitle    = {Proceedings of the 2025 Conference on Empirical Methods in Natural
                  Language Processing, {EMNLP} 2025, Suzhou, China, November 4-9, 2025},
  pages        = {32531--32548},
  publisher    = {Association for Computational Linguistics},
  year         = {2025},
  url          = {https://doi.org/10.18653/v1/2025.emnlp-main.1654},
  doi          = {10.18653/V1/2025.EMNLP-MAIN.1654},
  bibsource    = {dblp computer science bibliography, https://dblp.org}
}

@article{mtts10,
  author       = {Changjiang Gao and
                  Xu Huang and
                  Wenhao Zhu and
                  Shujian Huang and
                  Lei Li and
                  Fei Yuan},
  title        = {Could Thinking Multilingually Empower {LLM} Reasoning?},
  journal      = {CoRR},
  volume       = {abs/2504.11833},
  year         = {2025},
  url          = {https://doi.org/10.48550/arXiv.2504.11833},
  doi          = {10.48550/ARXIV.2504.11833},
  eprinttype    = {arXiv},
  eprint       = {2504.11833},
  bibsource    = {dblp computer science bibliography, https://dblp.org}
}

@inproceedings{mtts11,
  author       = {Sanchit Ahuja and
                  Praneetha Vaddamanu and
                  Barun Patra},
  editor       = {Christos Christodoulopoulos and
                  Tanmoy Chakraborty and
                  Carolyn Rose and
                  Violet Peng},
  title        = {EfficientXLang: Towards Improving Token Efficiency Through Cross-Lingual
                  Reasoning},
  booktitle    = {Findings of the Association for Computational Linguistics: {EMNLP}
                  2025, Suzhou, China, November 4-9, 2025},
  pages        = {15612--15624},
  publisher    = {Association for Computational Linguistics},
  year         = {2025},
  url          = {https://aclanthology.org/2025.findings-emnlp.845/},
  bibsource    = {dblp computer science bibliography, https://dblp.org}
}

@article{mtts12,
  author       = {Kang Chen and
                  Mengdi Zhang and
                  Yixin Cao},
  title        = {Less Data Less Tokens: Multilingual Unification Learning for Efficient Test-Time Reasoning in LLMs},
  journal      = {CoRR},
  volume       = {abs/2506.18341},
  year         = {2025},
  url          = {https://doi.org/10.48550/arXiv.2506.18341},
  doi          = {10.48550/ARXIV.2506.18341},
  eprinttype    = {arXiv},
  eprint       = {2506.18341},
}

@article{decoding,
  author       = {Max Peeperkorn and
                  Tom Kouwenhoven and
                  Dan Brown and
                  Anna Jordanous},
  title        = {Mind the Gap: Conformative Decoding to Improve Output Diversity of
                  Instruction-Tuned Large Language Models},
  journal      = {CoRR},
  volume       = {abs/2507.20956},
  year         = {2025},
  url          = {https://doi.org/10.48550/arXiv.2507.20956},
  doi          = {10.48550/ARXIV.2507.20956},
  eprinttype    = {arXiv},
  eprint       = {2507.20956},
  bibsource    = {dblp computer science bibliography, https://dblp.org}
}

\appendix

\section{Appendix}
\label{sec:appendix}

\begin{figure}[t]
    \centering
    \includegraphics[width=1.0\linewidth]{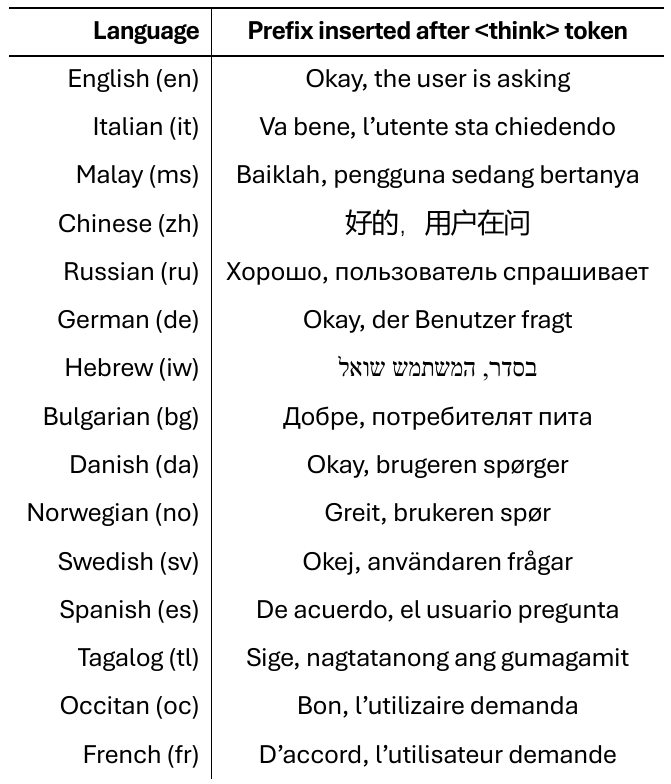}
    \caption{Prefix translations used for Thinking Language Control.}
    \label{fig:prefix}
\end{figure}

\subsection{Language Control Details}
\label{sec:lang_control_details}

\begin{table}[t]
\centering
\resizebox{\columnwidth}{!}{%
\begin{tabular}{llcc}
\toprule
Model & Lang & Think-Target (\%) & Output-EN (\%) \\
\midrule
\multirow{2}{*}{Qwen3-8B}
  & en      & 100.00 & 98.29 \\
  & non-en & 99.88 $\pm$ 0.25 & 98.28 $\pm$ 1.31 \\
\cmidrule(lr){1-4}
\multirow{2}{*}{Qwen3-14B}
  & en      & 100.00 & 98.37 \\
  & non-en & 99.57 $\pm$ 1.45 & 99.50 $\pm$ 0.35 \\
\cmidrule(lr){1-4}
\multirow{2}{*}{Qwen3-32B}
  & en      & 100.00 & 100.00 \\
  & non-en & 99.54 $\pm$ 1.47 & 98.61 $\pm$ 0.69 \\
\cmidrule(lr){1-4}
\multirow{2}{*}{DeepSeek-14B}
  & en      & 100.00 & 96.10 \\
  & non-en & 98.70 $\pm$ 2.57 & 95.32 $\pm$ 1.51 \\
\bottomrule
\end{tabular}}
\caption{
Sanity-check verification of thinking and output language control.
Results for English thinking are reported individually, while results for non-English thinking
are averaged over multiple languages and reported as mean $\pm$ standard deviation.
}
\label{tab:language_verification}
\end{table}

Figure~\ref{fig:prefix} presents the translated prefixes used for Thinking Language Control across 15 languages.
By inserting the corresponding prefix immediately after the \texttt{<think>} token,
the model is guided to conduct its intermediate thinking in the target language.

Combined with Output Language Control, the model is guided to thinking in a specified language
while producing English responses.
As a sanity check, we apply an off-the-shelf language identification tool\footnote{https://github.com/pemistahl/lingua-py}
to the thinking content within the \texttt{<think>}~\ldots~\texttt{</think>} span,
as well as to the final output following \texttt{</think>}.

Table~\ref{tab:language_verification} summarizes the averaged results on
\textsc{NoveltyBench} and \textsc{Infinity-Chat}.
Across models, the thinking segments are predominantly detected as the target thinking language,
and the output segments are predominantly detected as English.
Although language identification may introduce some noise,
these results indicate that the intended language control signals are largely reflected in the generated text.

\begin{table}[t]
\centering
\small
\begin{tabular}{p{0.95\columnwidth}}
\toprule
\textbf{Output Quality Evaluation Prompt} \\
\midrule
You are an evaluator assessing the quality of a single response to a task instruction. \\[0.5em]

You will be given: \\
(1) A task instruction \\
(2) A response \\[0.5em]

Evaluate the response along the following two dimensions: \\[0.5em]

1. Instruction Adherence (0--50) \\
To what extent does the response follow the task instruction? \\
Note that if the response explicitly refuses to perform the task, this should {NOT} be penalized. \\
You only need to judge the degree to which the response is relevant to the task instruction. \\[1em]

2. Response Quality (0--50) \\
Assess the overall quality of the response in terms of clarity, fluency, and grammatical correctness. \\[0.5em]

{Scoring:} \\
- Each dimension should be scored from 0 to 50 (integer only). \\
- Total Score = sum of the two dimensions (0--100). \\[0.5em]

{Output format (strict JSON only):} \\[0.3em]
\{ \\
\ \ \ "Instruction Adherence": <score>, \\
\ \ \ "Response Quality": <score>, \\
\ \ \ "Total Score": <score> \\
\} \\
\bottomrule
\end{tabular}
\caption{Prompt template used for output quality evaluation with \texttt{gpt-4o-mini}.}
\label{tab:quality_prompt}
\end{table}

\subsection{Output Quality Evaluation Details}
\label{sec:quality_evaluation}

Table~\ref{tab:quality_prompt} shows the complete prompt used for output quality evaluation.
The total quality score is computed as the sum of the two evaluation dimensions.
For each task instance, all sampled responses are evaluated independently,
and we report the average quality score across samples.

\begin{table*}[t]
\centering
\small
\begin{adjustbox}{max width=\textwidth}
\begin{tabular}{lc|cccccccccccccc|c}
\toprule
 & en & it & ms & zh & ru & de & iw & bg & da & no & sv & es & tl & oc & fr & avg (non-en) \\
\midrule
\multicolumn{17}{c}{\textit{Distinct Score $\uparrow$}} \\
\midrule
Qwen3-8B &
20.67 & 21.89 & 22.15 & \worst{20.13} & 20.47 & 22.87 &
23.98 & 23.64 & 23.10 & 24.51 & 22.65 & 20.73 &
23.71 & \best{24.47} & 21.27 & 22.54 \\

Qwen3-14B &
20.40 & 22.40 & 20.88 & 21.93 & 21.53 & 22.40 &
\best{27.07} & 21.47 & 23.67 & 24.47 & 22.80 & 21.00 &
23.85 & 23.23 & \worst{19.73} & 22.60 \\

Qwen3-32B &
27.00 & 27.60 & 27.67 & 27.20 & \worst{25.73} & 26.27 &
27.05 & 26.07 & 28.60 & 27.78 & 28.47 & 28.47 &
\best{28.66} & \best{28.66} & 27.00 & 27.52 \\

DeepSeek-14B &
\worst{25.27} & 30.53 & 29.00 & 28.80 & 29.33 & 30.33 &
35.76 & 30.88 & 34.40 & 34.00 & 35.20 & 27.93 &
\best{39.61} & 31.99 & 28.00 & 31.84 \\

\midrule
\multicolumn{17}{c}{\textit{Similarity Score $\downarrow$}} \\
\midrule
Qwen3-8B &
89.05 & 88.69 & 88.80 & 88.80 & \worst{89.30} & 87.83 &
87.36 & 88.09 & 88.12 & 87.47 & 88.30 & 88.75 &
88.26 & \best{86.78} & 88.64 & 88.23 \\

Qwen3-14B &
89.53 & 88.89 & 89.13 & 88.50 & 89.36 & 89.12 &
\best{87.77} & 88.83 & 88.53 & 88.18 & 88.60 & 89.36 &
88.81 & 88.37 & \worst{89.58} & 88.79 \\

Qwen3-32B &
85.24 & 81.97 & 84.98 & 82.89 & 84.27 & \best{76.49} &
\worst{86.22} & 85.52 & 82.54 & 84.10 & 79.24 & 80.83 &
85.72 & 83.77 & 82.31 & 82.92 \\

DeepSeek-14B &
\worst{85.97} & 83.16 & 85.52 & 85.74 & 84.09 & 83.06 &
\best{79.11} & 83.31 & 80.85 & 80.15 & 82.64 & 85.46 &
79.30 & 83.11 & 85.19 & 82.91 \\

\midrule
\multicolumn{17}{c}{\textit{Output Quality $\uparrow$}} \\
\midrule
Qwen3-8B &
\best{96.82} & 95.86 & 95.72 & 95.53 & 96.11 & 96.69 &
95.53 & 96.04 & \worst{95.09} & 95.00 & \best{96.82} & 95.72 &
95.70 & 95.59 & 95.40 & 95.77 \\

Qwen3-14B &
\best{96.93} & 94.94 & 95.48 & 95.03 & \worst{94.70} & 96.03 &
96.50 & 96.00 & 96.10 & 96.78 & 96.16 & 95.79 &
95.49 & 95.87 & 95.75 & 95.76 \\

Qwen3-32B &
\best{97.36} & 96.08 & 95.85 & 96.22 & 95.36 & 94.47 &
95.57 & 97.07 & 95.52 & 96.87 & 95.96 & 94.97 &
96.04 & 96.19 & \worst{94.26} & 95.74 \\

DeepSeek-14B &
88.46 & \best{89.45} & 88.99 & 89.44 & 90.71 & 86.79 &
86.51 & \worst{80.12} & 87.24 & 82.13 & 85.06 & 87.52 &
87.13 & 83.99 & 90.07 & 86.80 \\

\bottomrule
\end{tabular}
\end{adjustbox}
\caption{
Distinct Score (\%), Similarity Score (\%), and Output Quality across models and thinking languages under \textit{Single-Language Sampling} on \textsc{Infinity-Chat}.
For each row, the best and worst language results are highlighted.
}
\label{tab:diversity_infinity}
\end{table*}

\begin{figure}[t]
    \centering
    \includegraphics[width=1.0\linewidth]{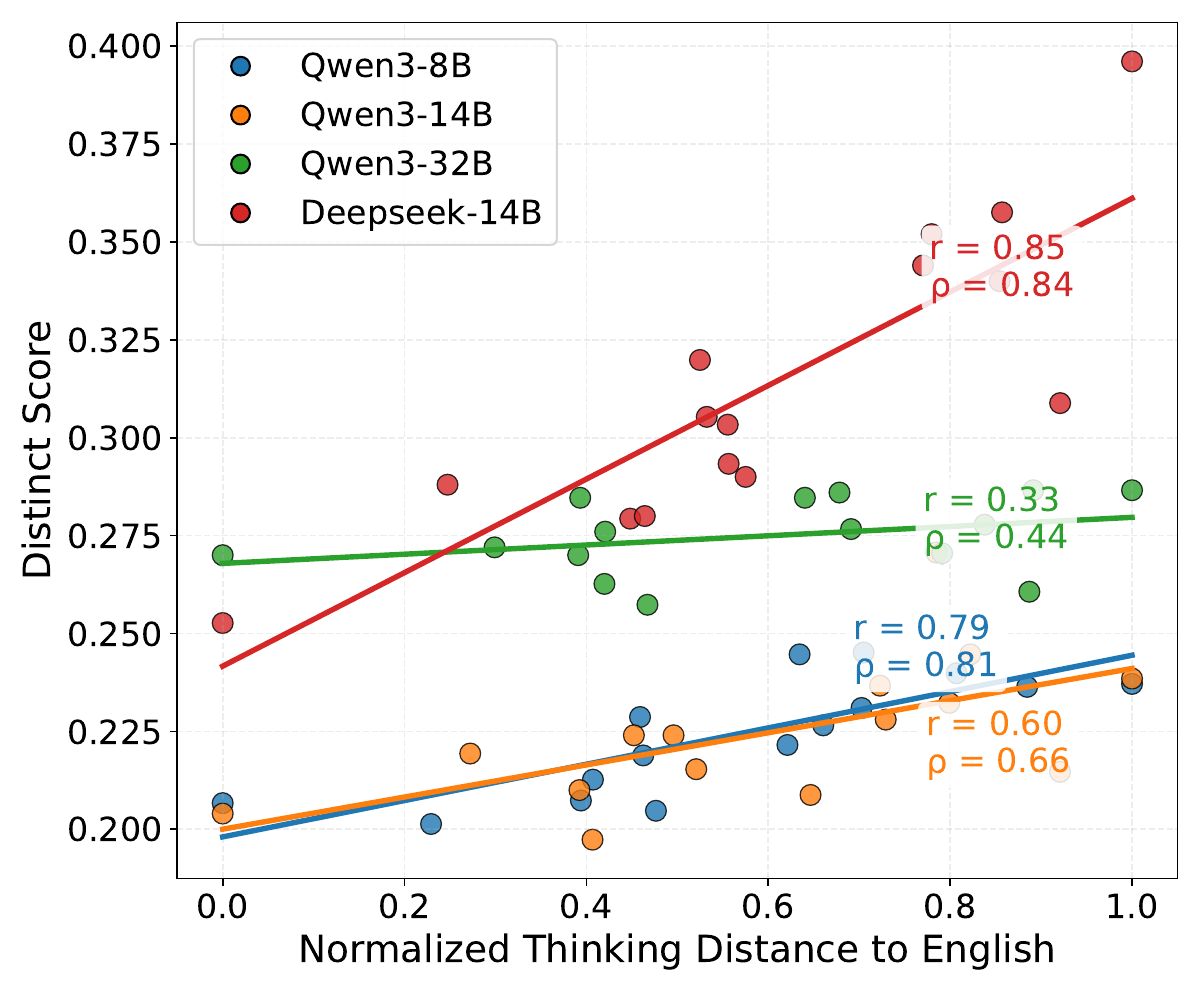}
    \caption{
    Correlation between the Distinct Score and the thinking distance to English across languages.
    Pearson’s $r$ and Spearman’s $\rho$ are reported for each model.
    Distinct Scores are obtained under \textit{Single-Language Sampling} on \textsc{Infinity-Chat}.
    Thinking distances are normalized to the range $[0,1]$ for visualization.
    }
    \label{fig:col_infi}
\end{figure}

\subsection{Additional Results on Single-Language Sampling}
\label{sec:res_infi}

Table~\ref{tab:diversity_infinity} reports the results of \textit{Single-Language Sampling}
on \textsc{Infinity-Chat}.
Overall, we observe several consistent trends that align with the main findings.
First, switching the language of thought from English to non-English languages
generally leads to higher output diversity across models,
as reflected by higher \textit{Distinct Score} and lower \textit{Similarity Score}.
Second, there exists notable variation across thinking languages:
languages such as \texttt{en}, \texttt{ru}, and \texttt{fr} tend to exhibit lower diversity,
whereas others, including \texttt{iw}, \texttt{tl}, and \texttt{oc},
consistently achieve higher diversity.
Finally, we do not observe a clear or systematic trade-off between output diversity and quality across languages.
Several non-English languages achieve improved diversity while maintaining
comparable output quality.

Figure~\ref{fig:col_infi} further reports the correlation between output diversity
and the thinking distance to English across languages on \textsc{Infinity-Chat}.
Consistent with our main results, we observe a strong positive correlation
for most models.
This result further corroborates that repeated sampling within thinking regions
farther from English is associated with higher output diversity.

\begin{table}[t]
\centering
\footnotesize
\setlength{\tabcolsep}{5pt}
\renewcommand{\arraystretch}{1.05}
\resizebox{\columnwidth}{!}{%
\begin{tabular}{lcccc}
\toprule
Model & S-en & S-non-en avg & S-best & Mixed \\
\midrule
\multicolumn{5}{c}{\textsc{NoveltyBench}} \\
\midrule
Qwen3-8B   & 87.28 & 84.72 & \textbf{80.79} & 82.84 \\
Qwen3-14B  & 87.82 & 86.78 & \textbf{85.04} & 85.29 \\
Qwen3-32B  & 82.10 & 79.99 & \textbf{77.65} & 79.44 \\
DeepSeek-14B
           & 81.15 & 79.86 & \textbf{76.16} & 77.64 \\
\midrule
\multicolumn{5}{c}{\textsc{Infinity-Chat}} \\
\midrule
Qwen3-8B   & 89.05 & 88.23 & 86.78 & \textbf{86.47} \\
Qwen3-14B  & 89.53 & 88.79 & \textbf{87.77} & 87.87 \\
Qwen3-32B  & 85.24 & 82.92 & \textbf{76.49} & 80.29 \\
DeepSeek-14B
           & 85.97 & 82.91 & \textbf{79.11} & 82.15 \\
\bottomrule
\end{tabular}%
}
\caption{
Similarity score (\%) comparison of \textit{Mixed-Language Sampling} and
\textit{Single-Language Sampling} on \textsc{NoveltyBench} and \textsc{Infinity-Chat}.
\textbf{Bold} indicates the best-performing sampling setting for each model and benchmark.
}
\label{tab:mixed_vs_single_similarity}
\end{table}

\subsection{Additional Results on Mixed-Language Sampling}
\label{sec:mix_result}

Table~\ref{tab:mixed_vs_single_similarity} compares \textit{Mixed-Language Sampling}
with three \textit{Single-Language Sampling} settings using the \emph{Similarity Score}.
Consistent with the main results, \textit{Mixed-Language Sampling} consistently outperforms
S-en and S-non-en avg, and in several cases matches or exceeds the S-best setting.
This shows that its advantage lies in improving diversity without requiring the selection
of a single best-performing language.

\subsection{Culture Evaluation Details}
\label{sec:culture_evaluation}

\paragraph{Datasets}
For \textsc{Blend}, we extract the set of unique questions from the original large-scale dataset
and merge all answer options into each question,
resulting in a multiple-choice dataset with 402 questions.
For \textsc{WVS}, the original dataset contains 290 questions.
We remove 8 questions without predefined options, yielding a final set of 282 multiple-choice questions.

\paragraph{Evaluation Protocols}
In \textsc{Blend}, each answer option is associated with one or more countries.
For each sampled response, we extract the selected option and increment the count of its associated country (or countries).
Let $p(c)$ denote the empirical distribution over countries aggregated from $M$ samples.
Cultural pluralism is measured as the normalized entropy:
\[
H_{\text{Blend}} = \frac{-\sum_{c} p(c)\log p(c)}{\log |C|}
\]
where $C$ denotes the set of all countries appearing in the answer options for the question.
The reported results are averaged over all questions.

In \textsc{WVS}, each sampled response corresponds to a discrete value option.
Let $p(o)$ denote the empirical distribution over predicted options across $M$ samples.
Cultural pluralism is defined as the normalized entropy:
\[
H_{\text{WVS}} = \frac{-\sum_{o} p(o)\log p(o)}{\log |O|}
\]
where $O$ denotes the set of possible value options for the question.
The reported results are averaged over all questions.

\paragraph{Baselines}
The \textit{Request Diversity} baseline appends the following sentence to the original instruction:
\emph{``Please try to provide a novel answer.''}

For \textit{Multilingual Prompting}, we use Google Translate to translate each original question from English into the same set of 14 non-English languages used in the main experiments.

\end{document}